\documentclass[11pt,a4paper]{article}
\usepackage[hyperref]{acl2021}
\usepackage{times}
\usepackage{latexsym}

\usepackage{microtype}

\aclfinalcopy 

\usepackage{latexsym}
\usepackage{tabularx}
\usepackage{graphicx}
\usepackage{float}
\usepackage{enumerate}
\usepackage{amssymb,amsmath}
\usepackage{color}
\usepackage{bm}
\usepackage{makecell}
\usepackage{multi row}
\usepackage{array}
\usepackage{resizegather}
\usepackage{refcount}
\usepackage{fixfoot}
\usepackage{hyperref}
\usepackage{url}
\usepackage{subfigure}  
\usepackage{bbm}
\usepackage{listings}

\usepackage{booktabs}
\usepackage{bm}
\usepackage{times}
\usepackage{latexsym}
\usepackage{amsmath,amsfonts,amssymb,bbm,epsfig,bm}
\usepackage{algpseudocode}
\usepackage{balance}
\usepackage{booktabs}
\usepackage{multirow}
\usepackage{array}
\usepackage{url}
\usepackage{tikz}
\usepackage{makecell}
\usepackage{pifont}

\newcommand{\red}[1]{{\color{black}#1}}

\definecolor{mypurple}{RGB}{153, 51, 255}

\newif\ifcomment
\commenttrue
\ifcomment

\newcommand{\veconospace}{\textsc{VeCo}}{}
\newcommand{\veco}{\textsc{VeCo~}}{}
\newcommand{\camlm}{\textsc{CA-MLM~}}

\newcommand{\plugin}{$_{{in}}$}
\newcommand{\plugout}{$_{{out}}$}
\newcommand{\vecoplugin}{\textsc{VeCo}\plugin}
\newcommand{\vecoplugout}{\textsc{VeCo}\plugout}

\bibpunct[, ]{(}{)}{;}{a}{,}{,}

\title{PlugIn\&PlugOut: Variable Encoder-decoder Pre-training for Cross-lingual Understanding and Generation}
\title{Plug-and-Play Cross-lingual Models for Language Understanding and Generation}
\title{Plug-and-Play Encoder-decoder Pre-training for Cross-lingual Understanding and Generation}
\title{Plug-and-Play Encoder-decoder Pre-training for Cross-lingual Understanding and Generation}
\title{VECO: Variable and Flexible Cross-lingual Pre-training for Language Understanding and Generation}

\author{Fuli Luo$^*$, Wei Wang$^*$, Jiahao Liu, Yijia Liu, Bin Bi, Songfang Huang, Fei Huang, Luo Si \\ Alibaba Group \\ 
{\tt\{lfl259702,hebian.ww,glacier.ljh,yanshan.lyj\}@alibaba-inc.com } \\
{\tt\{b.bi,songfang.hsf,f.huang,luo.si\}@alibaba-inc.com }
}

\begin{document}

\maketitle
\renewcommand{\thefootnote}{\fnsymbol{footnote}}
\footnotetext[1]{Equal contribution.}
\renewcommand{\thefootnote}{\arabic{footnote}}

\begin{abstract}

Existing work in multilingual pretraining has demonstrated the potential of cross-lingual transferability by training a unified Transformer encoder for multiple languages.  However, much of this work only relies on the shared vocabulary and bilingual contexts to encourage the correlation across languages, which is loose and implicit for aligning the contextual representations between languages. In this paper, we plug a cross-attention module into the Transformer encoder to explicitly build the interdependence between languages. It can effectively avoid the degeneration of predicting masked words only conditioned on the context in its own language. More importantly, when fine-tuning on downstream tasks, the cross-attention module can be plugged in or out on-demand, thus naturally benefiting a wider range of cross-lingual tasks, from language understanding to generation. 

As a result, the proposed cross-lingual model delivers new state-of-the-art results on various cross-lingual understanding tasks of the XTREME benchmark, covering text classification, sequence labeling, question answering, and sentence retrieval. For cross-lingual generation tasks, it also outperforms all existing cross-lingual models and state-of-the-art Transformer variants on WMT14 English-to-German and English-to-French translation datasets, with gains of up to 1$\sim$2 BLEU.~\footnote{Code and model are available at \url{https://github.com/alibaba/AliceMind/tree/main/VECO}}
\end{abstract}

\section{Introduction}

\begin{figure}[htb]
\centering
\begin{center}
\subfigure[XLM (MLM + TLM)]
{
\begin{minipage}[t]{0.5\linewidth}
\centering
\includegraphics[width=\textwidth]{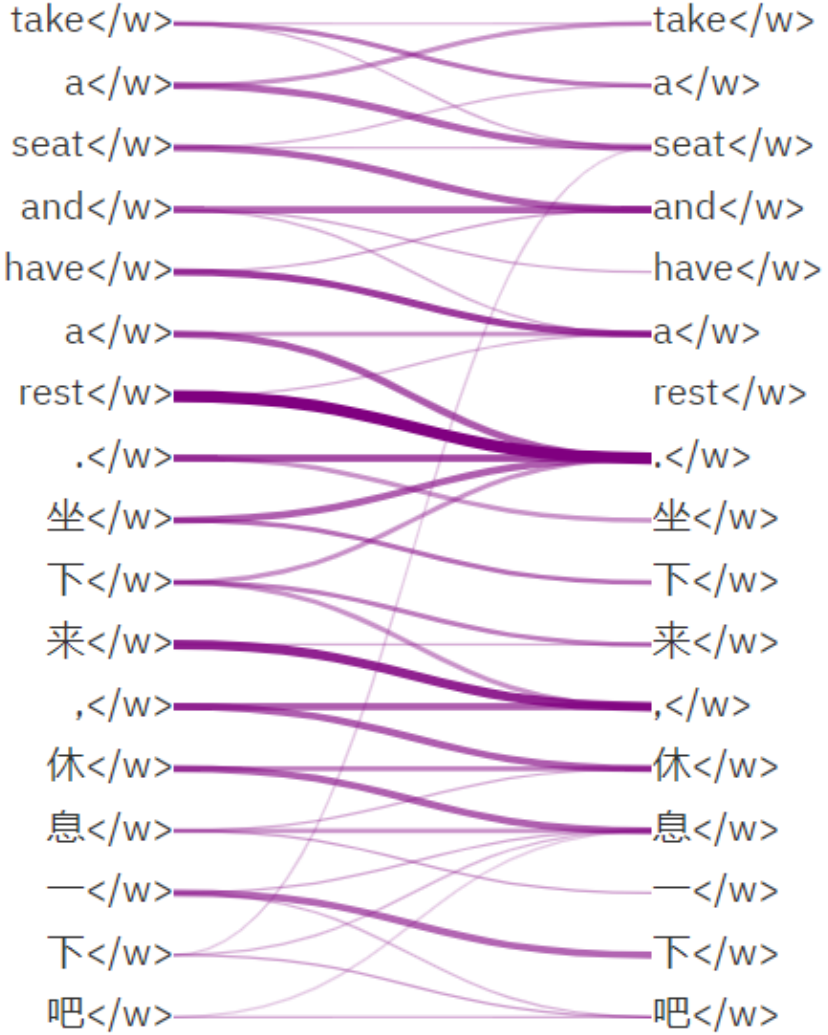}
\end{minipage}%
}%
\subfigure[XLM-R (MLM)]
{
\begin{minipage}[t]{0.45\linewidth}
\centering
\includegraphics[width=\textwidth]{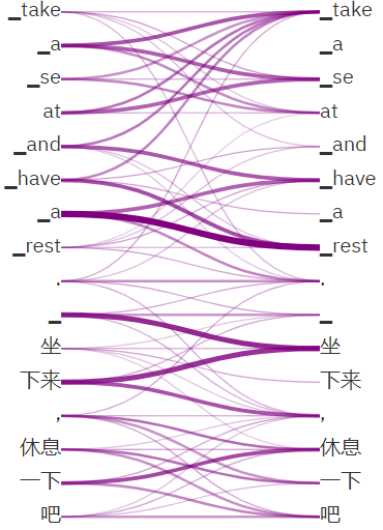}
\end{minipage}%
}%
\end{center}
\centering
\caption{The attention scores of XLM and XLM-R with the input of a pair of parallel sentences: \textit{Take a seat and have a rest} in English and its translated Chinese sentence. The darker line denotes a higher score. We can found that there are only a few attention patterns across English and Chinese subwords.}
\label{fig:motivation}
\end{figure}

\begin{figure*}[htb]
\centering
\begin{center}
\includegraphics[width=\textwidth]{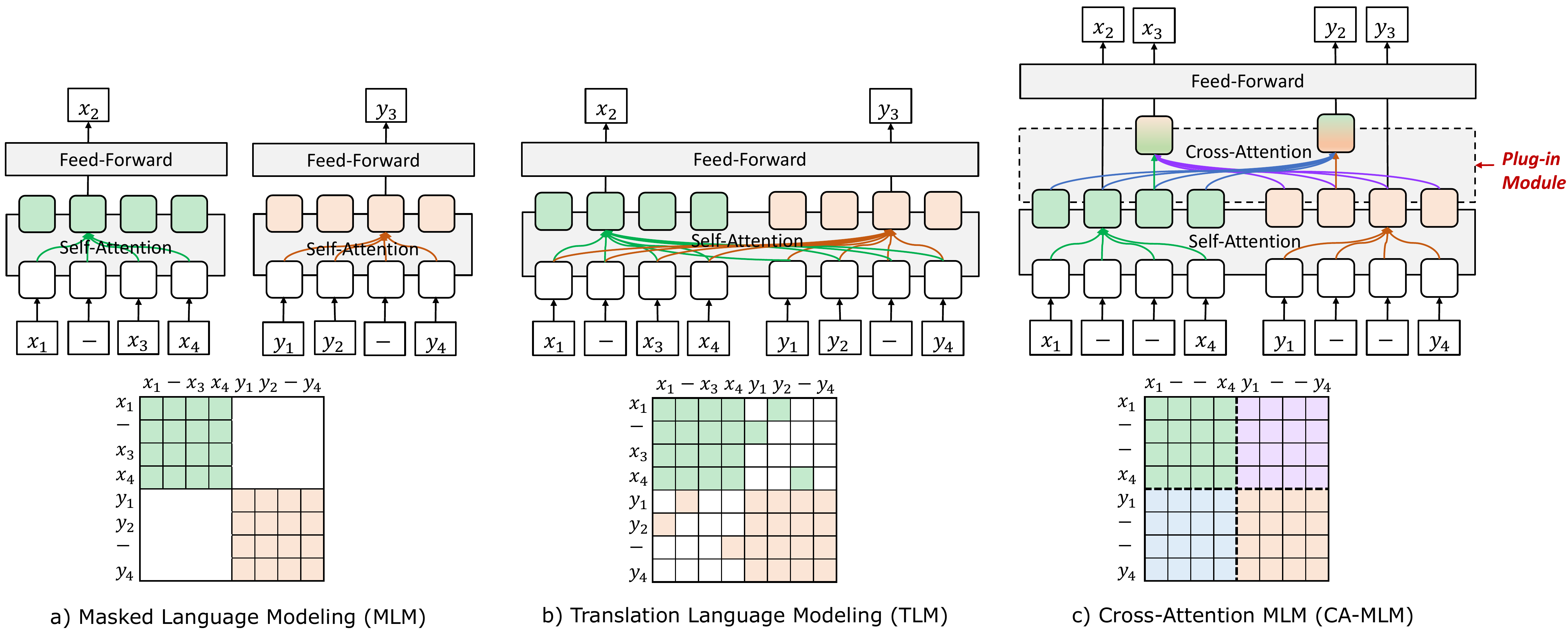}
\end{center}
\centering
\caption{A schematic comparison of cross-lingual pre-training tasks and their attention matrices. When predicting the masked words of different languages: a) MLM can only attend to the context in its own language; b) TLM implicitly attend to a part of words across languages (as shown in Figure~\ref{fig:motivation}). However, c) the proposed CA-MLM can: (1) not only attend to the context in its own language to predict words $\bm x_2$ and $\bm y_3$, (2) but also can firstly attend to its own context and then explicitly attend to all words across languages to predict words $\bm x_3$ and $\bm y_2$ via a plug-in cross-attention module.}
\label{fig:model}
\end{figure*}

Cross-lingual pre-trained models like mBERT~\cite{devlin2018bert}, XLM~\cite{conneau2019xlm} and XLM-R~\cite{conneau2019xlmr} that target providing contextualized representations for the inputs across languages, have shown large potential on a variety of cross-lingual understanding and generation tasks. 

Behind the great success, two major factors play the role of aligning the contextual representations between languages:
1) build the shared vocabulary across languages through subword tokenization, which supports the simple extension of masked language modeling (MLM) from English corpus to multilingual corpus; 2) capture the alignment in parallel data via concatenating two sentences as input, called translation language modeling (TLM). However, both of these two mechanisms rely on the \textbf{\textit{self-attention}} module (query=key/value) of the Transformer encoder to \textit{implicitly} enhance the interdependence between languages, which may lead to few attention patterns across languages. Taking Figure~\ref{fig:motivation} as an example, even though inputting a pair of parallel sentences, both models \textit{only} attend to the English context to build the representation of English tokens, while ignoring the semantically related Chinese tokens. That is, the self-attention module captures little communication across languages, which is crucial for learning universal cross-lingual representations.

Based on the above observation, we propose to plug a \textbf{\textit{cross-attention}} module (query!=key/value) into the Transformer encoder and design a cross-attention MLM task to \textit{explicitly} capture the interdependence between languages. As illustrated in Figure~\ref{fig:model} (c), the cross-attention module takes the representation of $\bm x$ as query and $\bm y$ as key/value (\textcolor{mypurple}{purple} lines) to build the representations of $\bm x$ in the next layer, thus explicitly aligning the representations across languages (purple attention matrices).
It can effectively avoid the degeneration of predicting masked words only conditioned on the context in its own language. 
Moreover, what distinguishes our work from pre-training an encoder-decoder model~\cite{liu2020mBART} is that we also keep the good nature (i.e., bidirectional contextual modeling) of the original encoder by \textit{unplugging} the cross-attention from the model to predicting the masked words (e.g., $\bm x_2$ and $\bm y_3$).  

Furthermore, when fine-tuning on various downstream tasks, we can choose either \textit{plug-in} or \textit{plug-out} the cross-attention module on-demand, thus making it suitable for both cross-lingual language understanding (NLU) and generation tasks (NLG). For cross-lingual NLU tasks, if plugging the cross-attention module out, we can adopt the same fine-tuning methods as an encoder-only model like XLM. However, we find that plugging the cross-attention module in fine-tuning can better utilize the bilingual context to boost the performance. For cross-lingual NLG like machine translation (MT), the cross attention is already jointly pre-trained with the whole network. Therefore, the parameters of the decoder do not need to be re-adjusted substantially in the following tuning process, thus fundamentally solving the main drawback of utilizing pre-trained encoders like XLM for initializing encoder-decoder models.

We call our approach \red{\veco for ``\underline{V}ariable and Fl\underline{e}xible \underline{C}r\underline{o}ss-lingual Pre-training''}.
We validate \veco on a variety of representative cross-lingual understanding and generation benchmarks.
Regrading cross-lingual understanding tasks, we conduct experiments on the XTREME benchmark consisting of 9 cross-lingual tasks, including text classification, sequence labeling, question answering, and sentence retrieval. \veco \red{ranks first at the XTREME leaderboard}~\footnote{\url{https://sites.research.google/xtreme}} at the submission deadline. 
Regrading cross-lingual generation tasks, we validate \veco on the widely used WMT14 English-German and English-French machine translation benchmarks. \veco obtains \red{44.5 and 31.7} BLEU scores, consistently outperforming existing cross-lingual pre-training approaches and state-of-the-art Transformer variants by around 1$\sim$2 BLEU.

\section{Pre-training of \veco}
\label{sec:method}

\subsection{Overview of \veco}
\veco extends from a multi-layer Transformer encoder and plugs a cross-attention module in each layer.
Given a pair of input $(\bm x, \bm y)$ and its corrupted version $(\hat{\bm x}, \hat{\bm y})$ via randomly masking part of its tokens, the model builds two types of contextualized vector representation for each token: 
\begin{itemize}
    \item One suit of contextual representations $\mathbf{H}$, denoted as green blocks and yellow blocks in Figure~\ref{fig:model} (c), are only build on self-attention module (i.e., unpluging the cross-attention module) in each layer. 
    \item Another suit of contextual representations $\mathbf{S}$, denoted as mixed color blocks in Figure~\ref{fig:model} (c), are build on both the self-attention and cross-attention modules~\footnote{For simplicity of illustration, we only show the mixed representations $\mathbf{S}$ of $\bm x_3$ and $\bm y_2$ in Figure~\ref{fig:model} (c).}. 
\end{itemize}

The model is trained to predict the masked tokens via two corresponding representations, conditioning on both its own context and paired context, respectively. Take predicting the masked words in sequence $\bm x$ as an example, the training objective is the cross-entropy of the gold distribution and predicted distribution $P({\bm x}|\hat{\bm x})$ and $P({\bm x}|\hat{\bm y}, \hat{\bm x})$ computed via the above two suits of contextual representations. Thus, the training objective of cross-attention masked language modeling (CA-MLM) can be formulated as
\begin{equation} \label{eq:total_loss}
\resizebox{0.42\textwidth}{!}{$%
\begin{split}
    & \mathcal{L}(\bm x,\bm y) = \\
    & -{\rm log}P({\bm x}|\hat{\bm x}; \mathbf{\theta_s}) -{\rm log}P({\bm x}|\hat{\bm y}, \hat{\bm x}; \mathbf{\theta_s},\mathbf{\theta_c}) \\
    & -{\rm log}P({\bm y}|\hat{\bm y}; \mathbf{\theta_s})  -{\rm log}P({\bm y}|\hat{\bm x}, \hat{\bm y}; \mathbf{\theta_s},\mathbf{\theta_c})
\end{split}$%
}
\end{equation}
where $\mathbf{\theta_s}$ and $\mathbf{\theta_c}$ are the parameters of self-attention and cross-attention modules.

\subsection{Architecture}
The backbone network of \veco is composed of a stack of $N$ Transformer layers. Each layer has three modules: a required self-attention module, a plug-and-play cross-attention module, and a required feed-forward linear module. Both self-attention and cross-attention modules are based on the multi-head attention~\cite{vaswani2017attention}. 
An attention function can be described as mapping a query ($\mathbf{Q}$) and a set of key-value ($\mathbf{K}$-$\mathbf{V}$) pairs to an output.

For the self-attention module, all the queries, keys and values are the same representations from the previous layer. Specifically, for the $l$-th Transformer layer, the output of a self-attention head $\mathbf{A}_l^s$ is computed via:
\begin{align}  \label{eq:self_attn1}
\mathbf{Q} &= \mathbf{H}^{l-1} \mathbf{W}_l^Q \\
\mathbf{K} &= \mathbf{H}^{l-1} \mathbf{W}_l^K \\ 
\mathbf{V} &= \mathbf{H}^{l-1} \mathbf{W}_l^V \\ 
\mathbf{A}_l^s &= \mathrm{softmax}(\frac{\mathbf{Q} \mathbf{K}^{\mathsf{T}}}{ \sqrt{d_k}}) \mathbf{V} \label{eq:self_attn2}
\end{align}
where $\mathbf{H}^{l-1}$ are the previous layer's outputs, $\mathbf{W}_l^Q , \mathbf{W}_l^K , \mathbf{W}_l^V$ are the parameter matrices of self-attention modules.

For the cross-attention module, the queries come from the previous layer, and the keys and values come from the last layer's representations of paired input. Specifically, for the $l$-th layer, the output of a cross-attention head $\mathbf{A}_l^c$ is computed via:
\begin{align} \label{eq:cross_attn1}
\mathbf{Q} &= \mathbf{S}^{l-1} \mathbf{U}_l^Q \\ 
\mathbf{K} &= \mathbf{H}^{L} \mathbf{U}_l^K \\
\mathbf{V} &= \mathbf{H}^{L} \mathbf{U}_l^V \\
\mathbf{A}_l^c &= \mathrm{softmax}(\frac{\mathbf{Q} \mathbf{K}^{\mathsf{T}}}{ \sqrt{d_k}}) \mathbf{V} \label{eq:cross_attn2}
\end{align}
where $\mathbf{S}^{l-1}$ are the previous layer's outputs, $\mathbf{U}_l^Q, \mathbf{U}_l^K, \mathbf{U}_l^V$ are the parameter matrices of cross-attention modules.

\begin{figure*}[htb]
\centering
\begin{center}
\includegraphics[width=0.9\textwidth]{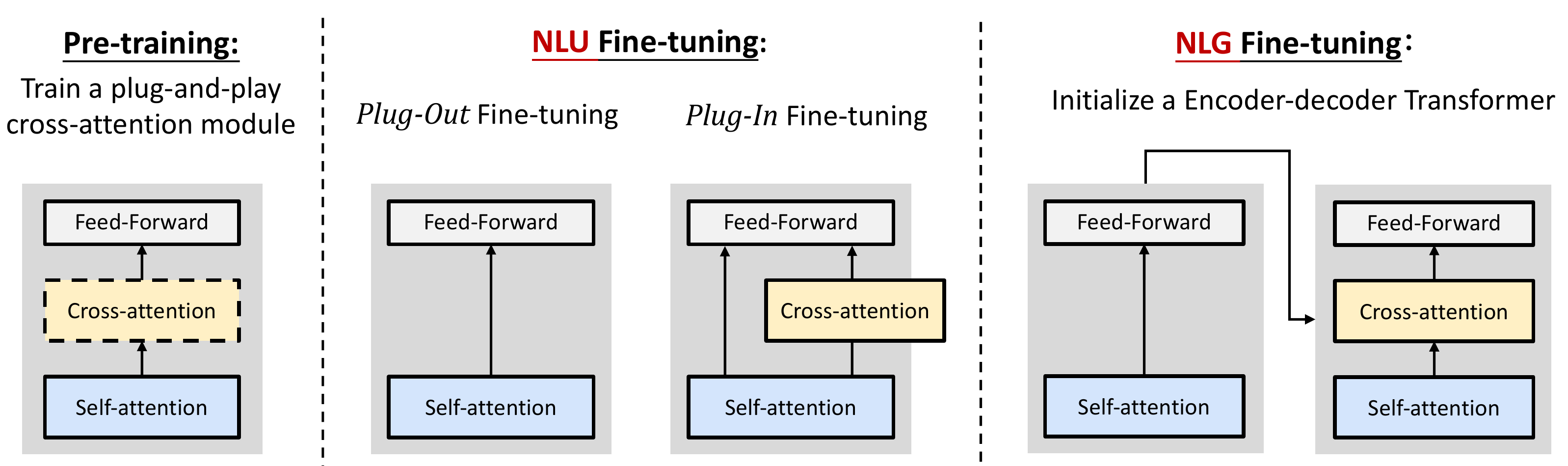}
\end{center}
\centering
\caption{The overview of \veconospace. During pre-training, a plug-and-play cross-attention module is jointly pre-trained along with the self-attention module. When fine-tuning on natural language understanding (NLU) tasks, the cross-attention module can be either plug-in or plug-out on demand.  When fine-tuning on natural language generation (NLG) tasks, \veco can initialize an encoder-decoder module (the mainstream backbone model of generation tasks) since all those necessary modules in the encoder and decoder are already pre-trained.}
\label{fig:model-pretraining-finetuning}
\end{figure*}

Finally, the output $\mathbf{H}^{L}$ of the last layer is used to recover the masked tokens of ${\bm x}$, conditioning on its own context. 
\begin{align} \label{eq:loss_1} 
    P({\bm x}|\hat{\bm x}) &= \mathrm{softmax}(f(\mathbf{H}^{L}_x)) \\
    P({\bm y}|\hat{\bm y}) &= \mathrm{softmax}(f(\mathbf{H}^{L}_y)) 
\end{align}
where $f$ is the feed-forward network that maps the output vectors into the dictionary. $\mathbf{H}^{L}_x$ and $\mathbf{H}^{L}_y$ are computed via Eq~\ref{eq:self_attn1}$\sim$\ref{eq:self_attn2} when $\mathbf{H}^{0}_x$ and $\mathbf{H}^{0}_y$ are the word embeddings of $\bm x$ and $\bm y$, respectively.

Meanwhile, $\mathbf{S}^{L}$, conditioning on the context of the paired sequence $\hat{\bm x}$ and $\hat{\bm y}$, is used to predict the masked tokens of ${\bm y}$.
\begin{align} \label{eq:loss_2}
    P({\bm x}|\hat{\bm y}, \hat{\bm x}) &= \mathrm{softmax}(f(\mathbf{S}^{L}_x)) \\
    P({\bm y}|\hat{\bm x}, \hat{\bm y}) &= \mathrm{softmax}(f(\mathbf{S}^{L}_y)) 
    \label{eq:loss_2_1}
\end{align}
where $\mathbf{S}^{L}_x$ and $\mathbf{S}^{L}_y$ are computed via Eq~\ref{eq:cross_attn1}$\sim$\ref{eq:cross_attn2} with the corresponding word embeddings and $\mathbf{H}^{L}$. 

Note that when optimizing the objectives based on Eq~\ref{eq:loss_2} and Eq~\ref{eq:loss_2_1}, we apply a stop-gradients operation~\cite{siamese_stop_gradients} to $\mathbf{H}^{L}$ (i.e., $\mathbf{H}^{L}$ is treated as a constant in this term). This operation can largely speed up the training by avoiding the backpropagation on a $2L$-layer network. Moreover, it even stabilizes the training of deep post-layernorm Transformer, which requires non-trivial efforts regarding carefully designing learning rate schedulers and cutting-edge optimizers~\cite{liu2020deeptransformer, bachlechner2020rezero}.

\section{Fine-tuning \veco for Downstream Cross-lingual Understanding and Generation Tasks}

As Figure~\ref{fig:model-pretraining-finetuning} illustrated, when fine-tuning on various downstream tasks, one advantage of \veco is its flexibility for initializing both the encoder-only Transformer for understanding tasks and encoder-decoder Transformer for generation tasks.  Beyond it, we also explore a fine-tuning approach combined with the characteristics of \veco.

\subsection{\veco for Cross-lingual Understanding}
Due to the plug-and-play cross-attention module, we explore two fine-tuning approaches:
\begin{itemize}
    \item \textit{Plug-Out} fine-tuning is to unplug the cross-attention module from the pre-trained model. In other words, the architecture of the fine-tuned model is almost the same as mBERT or XLM. Specifically, the contextual representations from the last layer $\mathbf{H}^{L}_x$ is used to predict the label of input $\bm x$.
   \item  \textit{Plug-In} fine-tuning is to plug the cross-attention module into the fine-tuned model, if the bilingual or automatically translated training data $\bm y$ is available in the downstream task. Specifically, we concatenated the two representations $[\mathbf{H}^{L}_x: \mathbf{S}^{L}_x]$ to predict the label of $\bm x$, $[\mathbf{H}^{L}_y: \mathbf{S}^{L}_y]$ to predict the label of $\bm y$.~\footnote{\textit{Plug-In} fine-tuning is not suitable for the zero-shot setting (also called \textit{cross-lingual transfer}) due to the lack of bilingual or translated pair $(\bm x, \bm y)$}.
\end{itemize}

\subsection{\veco for Cross-lingual Generation}

For pre-trained encoders like XLM, it is not a trivial problem to incorporate them into the sequence-to-sequence architecture -- the mainstream backbone model of generation tasks~\cite{IncorporatingBERT}.
One of the drawbacks or challenges could be that the encoder-to-decoder attention is not pre-trained. Therefore, the parameters of the decoder need to be re-adjusted along with the encoder in the following fine-tuning process~\cite{ren19explicit}. 

However, under the framework of \veco, the cross-attention is jointly pre-trained along with the whole network, making it easy to provide full initialization for sequence-to-sequence models. Specifically, the self-attention module is used to initialize both the corresponding modules in the encoder and decoder for contextual modeling, while the cross-attention module is used to initialize the encoder-to-decoder attention.
It's okay whether you continue to tie the self-attention parameters during fine-tuning.
Directly pre-training a sequence-to-sequence model like mBART~\cite{liu2020mBART} could be another solution for NLG tasks, but we found mBART is not so effective in cross-lingual NLU tasks. We refer the reader to the Section~\ref{sec:ablation} for detailed experiments and analysis.

\begin{table*}[t]
\centering
\resizebox{\linewidth}{!}{
\begin{tabular}{l|c|c|cc|c|c|c}
\toprule
Model & Architecture & \#Parameters & Enc Layers & Dec Layers & \#Languages &  \#Vocab & Training Data \\
\midrule
mBERT~\citep{devlin2018bert} & Encoder-only & 110M & 12~~ & - & 104 & 110k  & Wikipedia \\
XLM~\citep{conneau2019xlm} & Encoder-only & 570M & 24~~ & - & 100 & 200k  & Wikipedia \\
XLM-R~\citep{conneau2019xlmr}  & Encoder-only & 550M & 24~~ & - & 100 & 250k & CommonCrawl \\
mRASP~\cite{mRASP} & Encoder-decoder & 375M  & 6~~ & 6~~ & 32 & 64k & Translation \\
MMTE~\citep{siddhant2019mmte} & Encoder-decoder & 375M & 6~~ & 6~~ & 103 & 64k & Translation \\
mBART~\citep{liu2020mBART}  & Encoder-decoder & 680M & 12~~ & 12~~ & 25 & 250k & CommonCrawl \\ \midrule 
\veco   & Flexible & 662M & \multicolumn{2}{c|}{24*} & 50 & 250k & CommonCrawl + Translation \\ 
\bottomrule
\end{tabular}}
\caption{Comparison of large cross-lingual models. * denotes \veco unifies the encoder and decoder.}
\label{tab:comparison_details}
\end{table*}

\section{Pre-training Setup} \label{sec:pre_setup}

\paragraph{Model Configuration} We pre-train a 24-layer model with 1024 embedding/hidden size and 4096 feed-forward size. We do not use language embeddings to allow our model to better deal with downstream tasks of unseen languages. We adopt the same 250K vocabulary that is also used by XLM-R~\cite{conneau2019xlmr}. Table~\ref{tab:comparison_details} shows the other details of baselines and \veco.

\paragraph{Pre-Training Data}
We collect monolingual and bilingual corpus covering 50 languages.
For monolingual training datasets, we reconstruct CommonCrawl Corpus used in XLM-R~\cite{conneau2019xlmr}. We extract 1.36TB data in 50 languages, which contains 6.5G sentences and 0.4G documents. We up/down-sample the monolingual text like XLM from each language with a smoothing parameter $\alpha = 0.5$. For bilingual data, we collect from the OPUS website~\footnote{\url{http://opus.nlpl.eu/}} like previous works~\cite{conneau2019xlm, chi2020InfoXLM}. There are 6.4G parallel sentences, covering 879 language pairs across 50 languages. See more statistics of training data in Appendix A.

\paragraph{Optimization Settings}
For each iteration, we alternately sample a batch of adjacent segments from the monolingual corpus and a batch of parallel sentences from bilingual datasets to conduct a pair of masked input $(\hat{\bm x}, \hat{\bm y})$.
We adopt the translation language modeling (TLM) when the inputs are parallel bilingual sentences. Thus the overall training objective is the sum of TLM and the proposed CA-MLM objectives. During training, the model parameters except for cross-attention are initialized by XLM-R. We first freeze the parameters of XLM-R and only update the cross-attention parameters for faster convergence. Then, we jointly train the whole model. We pre-train our model with mixed-precision training using 64 Nvidia Telsa V100 32GB GPUs. Appendix A shows additional details.

\section{Experiments on Cross-lingual Understanding Tasks}
\begin{table*}[htp]
\small
\begin{center}
\begin{tabular}{lcccccccccc}
\toprule

 \textbf{Datasets} & \textbf{XNLI} & \textbf{PAWS-X} & \textbf{POS} & \textbf{NER} & \textbf{XQuAD} & \textbf{MLQA} & \textbf{TyDiQA} & \textbf{BUCC} & \textbf{Tatoeba} & \\
\#\textbf{Languages} & 15 & 7 & 33 & 40 & 11 & 7 & 9 & 5 & 33 & \\
\textbf{Metrics} & Acc & Acc & F1 & F1 & F1/EM & F1/EM & F1/EM & F1 & Acc & \bf Avg.  \\
\midrule 
\multicolumn{11}{l}{\textit{\textbf{Cross-lingual Transfer}: Fine-tune model on English training set and test on all languages}} \\
MMTE$^\dagger$ & 67.4 & 81.3 & 73.5 & 58.3 & 64.4/46.2 & 60.3/41.4 & 58.1/43.8 & 59.8 & 37.9 & 59.5 \\
mBERT$^\dagger$ & 65.4 & 81.9 & 70.3 & 62.2 & 64.5/49.4 & 61.4/44.2 & 59.7/43.0 & 56.7 & 38.7 & 59.6 \\
XLM$^\dagger$   & 69.1 & 80.9 & 70.1 & 61.2 & 59.8/44.3 & 48.5/32.6 & 43.6/29.1 & 56.8 & 32.6 & 55.5 \\
XLM-R$^\dagger$ & 79.2 & 86.4 & 72.6 & 65.4 & 76.6/60.8 & 71.6/\textbf{53.2} & 65.1/45.0 & 66.0 & 57.3 & 68.1 \\
\vecoplugout & \textbf{79.9} & \textbf{88.7} & \textbf{75.1} & \textbf{65.7} & \textbf{77.3/61.8} & \textbf{71.7/53.2} & \textbf{67.6/49.1} & \textbf{85.0} & \textbf{75.1} & \textbf{73.1} \\
\midrule
\multicolumn{11}{l}{\textit{\textbf{Translate-Train-All}: Fine-tune model on English training data and translated data of the target language}} \\
XLM-R$^\ddagger$ & 82.6 & 90.4 & - & - & 80.2/65.9 & 72.8/54.3 & 66.5/47.7 & - & - & -  \\  
XLM-R$^*$ & 82.8 & 90.2 & 72.6 & 65.4 & 80.0/65.8 & 73.0/54.3 & 74.5/58.3 & 80.2 & 75.2 & 74.4 \\ 
FILTER & 83.9 & 91.4 & 76.2 & 67.7 & 82.4/68.0 & 76.2/57.7 & 68.3/50.9 & 84.5 & 84.5 & 77.0 \\
\vecoplugout & 83.0 & 91.1 & 75.1 & 65.7 & 79.9/66.3 &  73.1/54.9 & 75.0/58.9 & 89.3 & 86.9  &  77.2 \\
\vecoplugin & \textbf{84.3} & \textbf{92.8} & \textbf{79.8} & \textbf{71.0} & \textbf{83.9/70.9} & \textbf{77.5/59.3} & \textbf{79.4/63.7} & \textbf{92.6} & \textbf{91.1} & \textbf{81.0} \\

\bottomrule
\end{tabular}
\caption{XTREME results on each dataset (as of ACL submission deadline). Averaged results on the four categories can be found at leaderboard: \url{https://sites.research.google/xtreme}. ``$^\dagger$'' and ``$^\ddagger$'' indicates results from~\citet{hu2020xtreme} and~\citet{fang2020filter}, respectively. ``*'' indicates the results obtained by our implementation.
The detailed results for each language are in Appendix \red{D}.
}
\label{tab:xtreme}
\vspace{-0.1in}
\end{center}
\end{table*}

\subsection{Experimental Setup}
\paragraph{Downstream Tasks}
We conduct cross-lingual NLU evaluations on XTREME~\cite{hu2020xtreme}, a representative massively multilingual benchmark that consists of 9 understanding tasks over 40 languages. XTREME tasks can be classified into four different categories: (1) sentence-pair classification:  XNLI~\cite{Conneau2018xnli}, PAWS-X~\cite{Yang2019paws-x}; (2) structured prediction: POS~\cite{nivre2018universal}, Wikiann NER~\cite{Pan2017}; (3) question answering: XQuAD~\cite{artetxe2020cross}, MLQA~\cite{Lewis2020mlqa}, TyDiQA~\cite{Clark2020tydiqa}; (4) sentence retrieval: BUCC 2018~\cite{zweigenbaum2018overview}, Tatoeba~\cite{Artetxe2019massively}.
Tasks in the first three categories are provided: 1) golden training corpus in English, 2) translated training corpus in other languages, and 3) dev/test set in all languages. For sentence retrieval tasks, no training datasets are provided.
We refer the reader to~\citet{hu2020xtreme} for additional details about the datasets.

\paragraph{Fine-tuning Setting}
Following previous works~\citep{conneau2019xlmr, hu2020xtreme}, we consider two typical fine-tuning settings: (1) \textit{Cross-lingual Transfer} which fine-tunes the pre-trained model using English golden data only and directly performs inference on the test data of different target languages; (2) \textit{Translate-Train-All} fine-tunes a multilingual model on the concatenation of all data (golden training corpus in English and translated training corpus in other languages). Note that for two sequence-labeling tasks (POS, NER), the position of token labels in the translated text generally differs from that in the source text. Following FILTER~\cite{fang2020filter}, we use the model trained only on the English training dataset as a teacher, to label the translated text.
To have a fair comparison with the strong baseline XLM-R~\cite{conneau2019xlmr} under the translate-train-all setting, we also show the results of XLM-R using the same fine-tuning hyperparameters as \veco.

\subsection{Experimental Results}

The detailed test results of nine tasks on the XTREME benchmark are shown in Table~\ref{tab:xtreme}. It demonstrates that the proposed \veco outperforms previous cross-lingual models on all datasets. 
Compared to XLM-R, it averagely scores \red{5.0 and 6.6} points higher under the cross-lingual transfer and translation-train-all settings, respectively.

In the cross-lingual transfer setting, \veco delivers a large improvement compared to XLM-R, especially on zero-shot sentence retrieval tasks (BUCC, Tatoeba). This phenomenon reflects that our model can better build the interdependence between languages. Thus it can better mine parallel sentences in a multilingual corpus.

Under the translation-train-all setting, it can be observed that \veco with Plug-In fine-tuning (\vecoplugin) is better than Plug-Out fine-tuning (\vecoplugout). We conclude the reasons as two-fold. On the input side, the Plug-Out fine-tuning individually takes multilingual instances as input, while the Plug-In fine-tuning considers the bilingual instances~\footnote{English instance with its translated one.} at each run. On the model side, the Plug-In fine-tuning can encourage correspondence across language via the cross-attention module. Note that the Plug-In fine-tuning method also outperforms FILTER~\cite{fang2020filter}, an enhanced cross-lingual fine-tuning method that also takes the bilingual instance as the input of XLM-R. It further demonstrates the effectiveness of \veco and its specialized fine-tuning method.

We conclude the reasons for the above performance improvement as two-fold: 1) the introduction of bilingual data during pre-training, which is a direct way to enhance the cross-lingual ability of the model; 2) Stronger ability to enhance the interdependence and fusion among languages via the proposed \camlm pre-training tasks. To analyze which plays a leading role, we conduct a set of more fair experiments in Section~\ref{sec:ablation}. 

\section{Experiments on Cross-lingual Generation Tasks}
\begin{table*}[]
\begin{minipage}[]{0.65\linewidth}
    \footnotesize
    \centering
    \begin{tabular}{lcccc}
      \toprule
        \multirow{2}{*}{\textbf{Model}} & \multicolumn{2}{c}{\textbf{WMT14 En-Fr}} & \multicolumn{2}{c}{\textbf{WMT14 En-De}} \\
        & BLEU & SacreBLEU & BLEU & SacreBLEU \\
        \midrule
        \multicolumn{5}{l}{\textit{Randomly Initialize}} \\
        Baseline  & 42.9 & 40.4 & 28.7 & 27.8 \\
        \citet{liu2020deeptransformer} & 43.8 & 41.8 & 30.1 & 29.5 \\
        \midrule
        \multicolumn{5}{l}{\textit{Randomly Initialize + More Bilingual Data}*} \\
        Baseline* & -  & - &  30.6  &  29.5 \\  
        \midrule
        \multicolumn{5}{l}{\textit{Cross-lingual Model Initialize}} \\
         mBART & 43.2 & 41.0 & 30.0 & 29.1 \\  
         \red{mRASP} & 44.3 & 41.7 & 30.3 & - \\  
         XLM-R & 43.8 & 41.2 & 30.9 & 29.9 \\
        \veco & \textbf{44.5} & \textbf{42.0} & \textbf{31.7} & \textbf{30.6} \\
        \bottomrule 
    \end{tabular}
\end{minipage}
\hfill
\begin{minipage}[]{0.35\linewidth}
     \vspace{9pt}
    \centering
    \resizebox{\columnwidth}{!}{
    \includegraphics{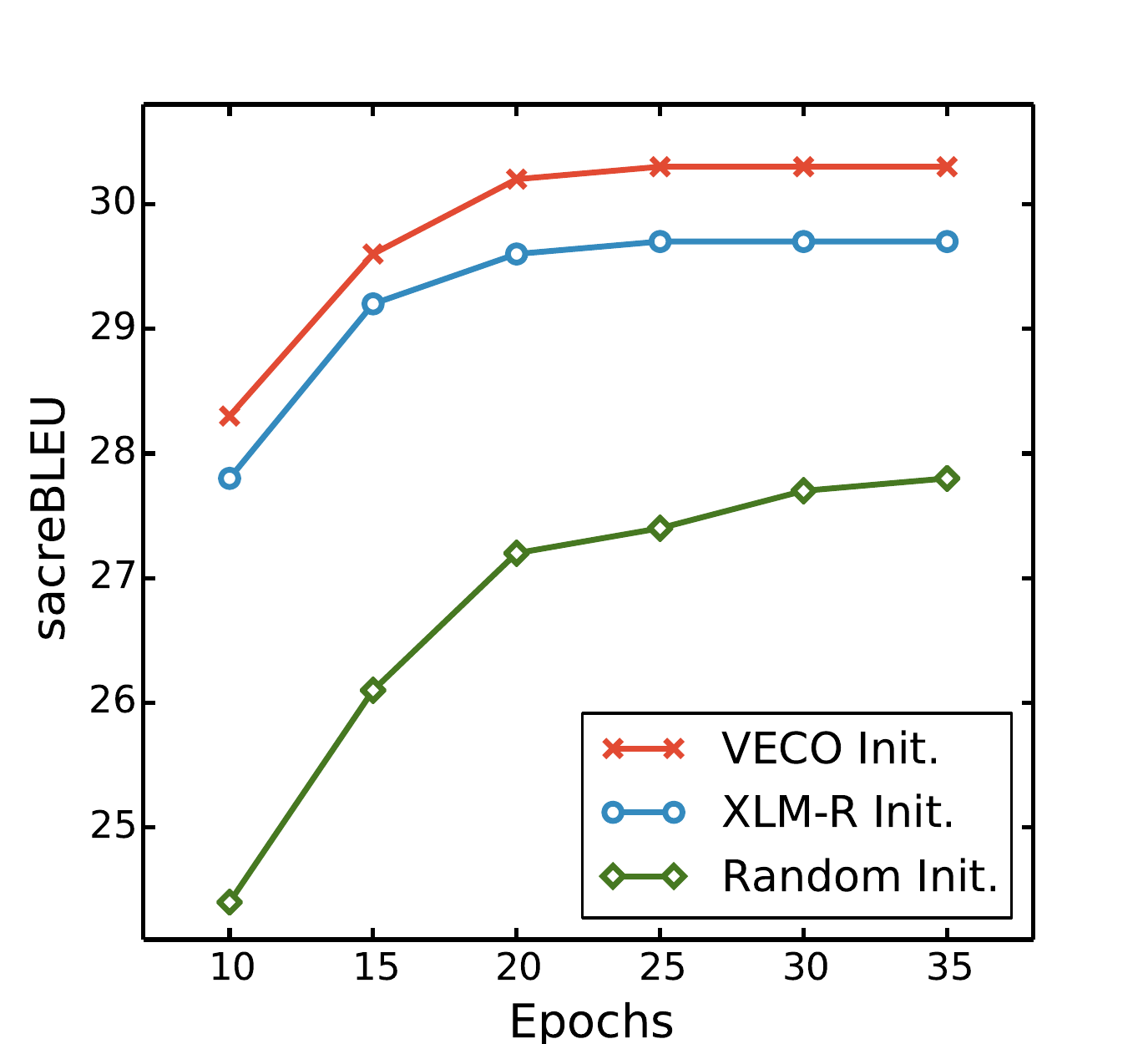}}
    \label{fig:ablation_pruning}
\end{minipage}
\hfill
\caption{(left) Results on machine translation. (right) Learning curves of different initialization methods.} \label{tb:mt_main_results}
\end{table*}

\subsection{Experimental Setup}
\paragraph{Datasets} 
We choose the machine translation (MT) task, a typical cross-lingual generation scenario. In order to illustrate the generality of our approach and have a fair comparison with the most recent state-of-the-art Transformer work~\cite{liu2020deeptransformer}, we choose two most widely used datasets: WMT14 English$\rightarrow$German (En-De) and English$\rightarrow$French (En-Fr) translation.
WMT14 En-De is a medium-resource dataset that provides 4.5M pairs for training and validation. We adopt standard newstest2014 as the test set. WMT14 En-Fr is a high-resource dataset that contains 36M pairs of parallel sentences. We use newstest2012+newstest2013 for validation and newstest2016 for test.
We measure case-insensitive tokenized BLEU with \texttt{multi-bleu.perl} and de-tokenized \texttt{SacreBLEU}~\footnote{Hash: BLEU+case.mixed+lang.en-\{de,fr\}+numrefs.1+\\smooth.exp+test.wmt14/full+tok.13a+version.1.4.9} to avoid the influence of different tokenization and normalization between models~\citep{Post18sacreBLEU}.

\paragraph{Fine-tuning Setting}
We fine-tune our model using \texttt{fairseq}~\footnote{\url{https://github.com/pytorch/fairseq}} toolkit and adopt comparable training settings with baselines.
We run WMT 14 En-De and En-Fr MT experiments on 16 and 32 V100 GPUs, respectively. The batch size is 64k for En-De and 256k for En-Fr. The total training updates are set to 100k. The learning rate is 1e-4/2e-4, with linear warm-up over the first 16k steps and linear decay.
We average the last 10 checkpoints and use beam search with a beam size of 5.

\paragraph{Baselines}
We consider two types of Transformer baselines: randomly initialized and cross-lingual models initialized. For random initialization, we reproduce a Transformer baseline that adopts the same architecture and fine-tuning hyperparameters as \veco but with random initialization. Besides, we compare to the state-of-the-art Deep Transformer~\cite{liu2020deeptransformer}.
For cross-lingual encoder-decoder models, we include mBART~\cite{liu2020mBART} and mRASP~\cite{mRASP}, which show impressive results on MT. Note that since we tied the self-attention weights of each encoder layer with each decoder layer, the whole parameters of mBART and \veco are comparable.
We also conduct the WMT experiments for XLM-R, following the totally same fine-tuning settings as \veco, but leaving the encoder-to-decoder attention un-initialized.

\subsection{Experimental Results} \label{sec:nlg_res}

Table~\ref{tb:mt_main_results} (left) shows the results on the machine translation.
We can observe that \veco can largely outperform the randomly initialized same-sized Transformer baseline by 2.3 BLEU points. Moreover, it even beats the (randomly initialized) state-of-the-art Deep-Transformer~\cite{liu2020deeptransformer}, which is three times deep as \veco.
Among the cross-lingual models, \veco can consistently outperform the best models, averaged on two datasets, by 0.8 BLEU points. 

Table~\ref{tb:mt_main_results} (right) displays the BLEU scores of same-sized models during training. We find that \veco initialized model can get a surprising more than 28 SacreBLEU score just after 10 epochs, which is better than the final score of the randomly initialized model at 35 epochs. It reveals that \veco can provide a fairly good initialization for the machine translation model, which can converge quickly and further boost the results.

One might suspect that the main reason for the performance improvement is leveraging parallel corpus during pre-training. To figure it out, we conduct a more comparable experiment. We first train an out-of-domain Transformer model using the whole En-De parallel data ($\sim$ 68M) used in \veco pre-training, and then continue to train the model on the in-domain WMT14 En-De training dataset. Results are shown in  Table~\ref{tb:mt_main_results} (left) marked with *. Under this set of a totally fair comparison, \veco still maintains a lead of 1.1 BLEU score. This directly confirms that the improvement in MT is not only due to the use of bilingual data. More importantly, \camlm ensures \textit{better} use of bilingual and large-scale unlabeled multilingual corpus.

\subsection{Potential of Initializing Shallow Decoder} \label{sec:nlg_res_more}
Online translation applications usually have a restriction of inference time. The most direct way is to reduce the decoder layers since previous MT works~\cite{liu2020deeptransformer} have shown that deeper encoders are more worthwhile than deeper decoders. Based on this, we also explore the potential of the \veco to initialize deep encoder and shallow decoder Transformers, which is a blank in the cross-lingual pre-training works.

Table~\ref{tb:mt_more_results} contrasts two ways of initializing a Transformer with $n$ decoder layers $(n<24)$ via selecting: (1) the first $n$ layers; (2) the last $n$ layers from a 24-layer pre-trained \veco model. We consider $n=\{3, 6\}$ to conduct experiments. We find that selecting the last $n$ layers exhibits better performance than selecting the first $n$ layers. It reveals that the last several layers play a more important role in making predictions over the whole vocabulary. Moreover, we can find that there is 0.2$\sim$0.3 BLEU gain when increasing the decoder layers from 3 to 6. However, we observe that only marginal improvement can be gained when further increasing the decoder layers from 6 to 24, which is also in line with the findings in~\citet{liu2020deeptransformer}.
Regardless of the initialization method, the \veco initialized model can gain consistent 1$\sim$2 BLEU improvement over the randomly initialized model.

\begin{table}[]
\footnotesize
\centering
\resizebox{\linewidth}{!}{
 \begin{tabular}{lccc}
  \toprule
    \multirow{2}{*}{\textbf{Method}} & \multirow{2}{*}{\textbf{\#Layers}}  & \multicolumn{2}{c}{\textbf{WMT14 En-De}} \\
    & & BLEU & SacreBLEU \\
    \midrule
    \multirow{2}{*}{\textit{Randomly Initialize}}
    & 3  & 28.5 & 27.6 \\ 
    & 6  &  28.6  &  27.7 \\
    \midrule
    \multirow{5}{*}{\textit{\veco Initialize}}
     & First-3 & 30.8 & 29.8 \\
     & Last-3  & 31.2 & 30.3 \\
     & First-6 & 31.1 & 30.1 \\
     & Last-6  & 31.5 & 30.5 \\
     & Full-24  & \textbf{31.7} & \textbf{30.6} \\
    \bottomrule
  \end{tabular}}
  \caption{Results of utilizing \veco to initialize deep encoder and shallow decoder (3/6-layer) Transformers.}
  \label{tb:mt_more_results}
\end{table}

\section{Analysis and Ablation Study} \label{sec:ablation}

We perform an ablation study to investigate where the improvement in cross-lingual NLU and NLG tasks mainly comes from. Specifically, there are three main aspects we have studied:
\begin{enumerate}
  \item How much performance improvement comes from the parallel translation corpus used in pre-training?
  \item How effective of the \camlm pre-training task, especially compared to the MLM and TLM pre-training tasks?
  \item How about pre-training a sequence-to-sequence model like mBART for NLU and NLG tasks?
\end{enumerate}

To figure out these questions, we train XLM, mBART and \veco model from scratch using the same datasets and parameter settings (see Appendix A for more details). All of them is pre-trained via MLM and TLM tasks. Note that the MLM task generally refers to predict the masked words of source language, while the TLM task generally refers to predict the words of the target language. Specifically for mBART that is under the framework of encoder-decoder, the input of encoder is masked sequence $ {\hat{\bm x}}$, and the target of decoder is the masked words of source input ${\bm x}$ (for MLM task), or the parallel sentence ${\bm y}$ (for TLM task).

Table~\ref{tb:ablation_results} shows the results of two representative datasets of cross-lingual NLU and NLG.
We can observe that, when using monolingual corpus only, \veco can outperform XLM by 0.8 points on the XNLI dataset and 0.3 BLEU scores on the IWSLT14 De-En translation dataset. It suggests that the \camlm can still benefit from adjacent sentences in monolingual corpus~\footnote{As noted in Section~\ref{sec:pre_setup}, we take two adjacent sentences in the monolingual corpus as $(\bm x, \bm y)$.}, to be equipped with a stronger ability of contextual modeling. Moreover, when pre-training both on the monolingual and bilingual corpus, \veco can even achieve a larger improvement compared to XLM, with 3.2 and 2.1 points improvement on two datasets, respectively. It reveals that \camlm objective of \veco can better utilize the bilingual corpus, compared to only optimized by TLM and MLM of XLM.

\begin{table}[]
\resizebox{0.5\textwidth}{!}{
\begin{tabular}{ccccc}
\toprule
\multirow{1}{*}{\textbf{Data}} & \multicolumn{1}{c}{\textbf{Models}} &  \multicolumn{1}{c}{\textbf{Tasks}} & \multicolumn{1}{c}{\textbf{XNLI}} & \multicolumn{1}{c}{\textbf{IWSLT}} \\
\midrule
\multirow{3}{*}{Mono.}
& XLM & MLM & 59.8 & 33.7 \\ 
& mBART & MLM & 57.3 & 32.9 \\
& \veco & \camlm & \textbf{60.6} & \textbf{34.0}  \\ \midrule
\multirow{3}{*}{Bili.}
& XLM  &  MLM+TLM & 64.5 & 33.9 \\ 
& mBART & MLM+TLM & 60.8 & 34.5 \\
& \veco & \camlm+TLM & \textbf{67.7} & \textbf{36.0} \\
\bottomrule
\end{tabular}}
\caption{Ablation study of small-sized models on XNLI and IWSLT14 De-En translation dataset.}
\label{tb:ablation_results}
\end{table}

Moreover, we find that pre-training a sequence-to-sequence model like mBART~\cite{liu2020mBART} performs worst on NLU tasks like XNLI~\footnote{We follow BART~\cite{lewis2019bart} by utilizing the final representation from the decoder for classification tasks.}, almost 6 points worse than \veco and near 2 points worse than XLM. One possible explanation could be that the unidirectional language modeling in the decoder might be sub-optimal for NLU tasks. And even on the machine translation task, mBART still performs worse than \veco when pre-training on the same bilingual datasets. 
We conclude that it is because that \veco can do better in the contextual modeling of source input $\bm x$ via a explicit masked language modeling objective in Eq~\ref{eq:loss_1} applied to $\bm x_2$ in Figure~\ref{fig:model} (c).

\section{Related Work}

mBERT~\cite{devlin2018bert} is a key step towards building a unified contextual language representation over multiple languages. It simply shares all languages' vocabulary and trains a bidirectional Transformer encoder, achieving promising results in various cross-lingual NLU tasks. There have been several extensions that follow the same encoder-only backbone as mBERT. The main difference is the introduction of more training corpus (e.g., bilingual data) and pre-training tasks. XLM~\cite{conneau2019xlm} utilizes both monolingual and bilingual corpus to perform the masked language modeling. XLM-R~\cite{conneau2019xlmr} extends to be built on RoBERTa~\cite{liu2019roberta} using larger monolingual training data. Other works~\cite{huang2019Unicoder, yang2020ALM, chi2020InfoXLM} propose new pre-training tasks to utilize the bilingual data better. However, there are two main drawbacks of these works. First, they mainly rely on the self-attention module in the Transformer encoder to implicitly build the interdependence between languages, leading to few attention patterns across languages due to the ``lazy'' network. Second, even though they show impressive performance improvement on cross-lingual understanding tasks like XNLI, only marginal improvement has been gained on cross-lingual generation tasks like machine translation, especially on high-resource languages.

A feasible solution for cross-language generation is to pre-train a denoising auto-encoder like mBART~\cite{liu2020mBART}. It extends BART~\cite{lewis2019bart} to the multilingual setting, demonstrating significant gains in low/medium-resource machine translation, but with a decrease in high resource languages. Unlike mBART, \citet{chi2020XNLG} first trains an encoder via MLM and then frozen the encoder to train the decoder only via two generative tasks. A similar approach is also proposed in \citet{liang2020xglue} and \citet{mRASP}, with the main difference in the joint training of encoder-decoder with code-switch tricks. However, all these cross-lingual models emphasize training a dedicated model for NLG. Thus they may hurt the NLU capabilities of the model. The ablation study in Section~\ref{sec:ablation} also validates that it is sub-optimal to train an encoder-encoder network for NLU tasks.

This paper endeavors to build a unified cross-lingual model for NLU and NLG tasks via a plug-and-play cross-attention module. More importantly, the cross-attention module plays a role in the explicit alignment of encoded representations of different languages, thus largely contributing to building a unified cross-lingual model.

\section{Conclusion}
We present \veconospace, a variable and flexible cross-lingual pre-training model, targets at explicitly capturing the interdependence between languages via a plug-and-play cross-attention module. Based on the flexible characteristics, \veco can initialize both NLU preferred encoder-only and NLG specialized encoder-decoder Transformer. Moreover, we also introduce a Plug-In fine-tuning approach to encourage the fusion between languages, combining the feature of \veco and cross-language downstream tasks.

Taken together, \veco achieves consistent improvements on various language understanding and generation tasks, \red{broadening the way of thinking about pre-trained backbone architecture and fine-tuning methods under the cross-lingual scenario.}

\bibliographystyle{acl_natbib}
\bibliography{anthology,acl2021}

\begin{thebibliography}{33}
\expandafter\ifx\csname natexlab\endcsname\relax\def\natexlab#1{#1}\fi

\bibitem[{Artetxe et~al.(2020)Artetxe, Ruder, and Yogatama}]{artetxe2020cross}
Mikel Artetxe, Sebastian Ruder, and Dani Yogatama. 2020.
\newblock \href {https://doi.org/10.18653/v1/2020.acl-main.421} {On the
  cross-lingual transferability of monolingual representations}.
\newblock In \emph{Proceedings of the 58th Annual Meeting of the Association
  for Computational Linguistics, {ACL} 2020, Online, July 5-10, 2020}, pages
  4623--4637. Association for Computational Linguistics.

\bibitem[{Artetxe and Schwenk(2019)}]{Artetxe2019massively}
Mikel Artetxe and Holger Schwenk. 2019.
\newblock \href {https://transacl.org/ojs/index.php/tacl/article/view/1742}
  {Massively multilingual sentence embeddings for zero-shot cross-lingual
  transfer and beyond}.
\newblock \emph{Trans. Assoc. Comput. Linguistics}, 7:597--610.

\bibitem[{Bachlechner et~al.(2020)Bachlechner, Majumder, Mao, Cottrell, and
  McAuley}]{bachlechner2020rezero}
Thomas Bachlechner, Bodhisattwa~Prasad Majumder, Huanru~Henry Mao, Garrison~W
  Cottrell, and Julian McAuley. 2020.
\newblock \href {https://arxiv.org/pdf/2003.04887.pdf} {Rezero is all you need:
  Fast convergence at large depth}.
\newblock \emph{arXiv preprint arXiv:2003.04887}.

\bibitem[{Chen and He(2020)}]{siamese_stop_gradients}
Xinlei Chen and Kaiming He. 2020.
\newblock \href {http://arxiv.org/abs/2011.10566} {Exploring simple siamese
  representation learning}.
\newblock \emph{CoRR}, abs/2011.10566.

\bibitem[{Chi et~al.(2020{\natexlab{a}})Chi, Dong, Wei, Wang, Mao, and
  Huang}]{chi2020XNLG}
Zewen Chi, Li~Dong, Furu Wei, Wenhui Wang, Xian-Ling Mao, and Heyan Huang.
  2020{\natexlab{a}}.
\newblock \href {https://arxiv.org/pdf/1909.10481.pdf} {Cross-lingual natural
  language generation via pre-training.}
\newblock In \emph{Proceedings of the AAAI Conference on Artificial
  Intelligence}.

\bibitem[{Chi et~al.(2020{\natexlab{b}})Chi, Dong, Wei, Yang, Singhal, Wang,
  Song, Mao, Huang, and Zhou}]{chi2020InfoXLM}
Zewen Chi, Li~Dong, Furu Wei, Nan Yang, Saksham Singhal, Wenhui Wang, Xia Song,
  Xian-Ling Mao, Heyan Huang, and Ming Zhou. 2020{\natexlab{b}}.
\newblock \href {https://arxiv.org/pdf/2007.07834.pdf} {{InfoXLM}: An
  information-theoretic framework for cross-lingual language model
  pre-training}.
\newblock \emph{arXiv preprint arXiv:2007.07834}.

\bibitem[{Clark et~al.(2020)Clark, Choi, Collins, Garrette, Kwiatkowski,
  Nikolaev, and Palomaki}]{Clark2020tydiqa}
Jonathan~H. Clark, Eunsol Choi, Michael Collins, Dan Garrette, Tom Kwiatkowski,
  Vitaly Nikolaev, and Jennimaria Palomaki. 2020.
\newblock {TyDi QA: A Benchmark for Information-Seeking Question Answering in
  Typologically Diverse Languages}.
\newblock In \emph{Transactions of the Association of Computational
  Linguistics}.

\bibitem[{Conneau et~al.(2019)Conneau, Khandelwal, Goyal, Chaudhary, Wenzek,
  Guzm{\'a}n, Grave, Ott, Zettlemoyer, and Stoyanov}]{conneau2019xlmr}
Alexis Conneau, Kartikay Khandelwal, Naman Goyal, Vishrav Chaudhary, Guillaume
  Wenzek, Francisco Guzm{\'a}n, Edouard Grave, Myle Ott, Luke Zettlemoyer, and
  Veselin Stoyanov. 2019.
\newblock \href {https://arxiv.org/pdf/1911.02116v1.pdf} {Unsupervised
  cross-lingual representation learning at scale}.
\newblock \emph{arXiv preprint arXiv:1911.02116}.

\bibitem[{Conneau et~al.(2018)Conneau, Rinott, Lample, Williams, Bowman,
  Schwenk, and Stoyanov}]{Conneau2018xnli}
Alexis Conneau, Ruty Rinott, Guillaume Lample, Adina Williams, Samuel Bowman,
  Holger Schwenk, and Veselin Stoyanov. 2018.
\newblock {XNLI}: Evaluating cross-lingual sentence representations.
\newblock In \emph{Proceedings of EMNLP 2018}, pages 2475--2485.

\bibitem[{Devlin et~al.(2019)Devlin, Chang, Lee, and
  Toutanova}]{devlin2018bert}
Jacob Devlin, Ming-Wei Chang, Kenton Lee, and Kristina Toutanova. 2019.
\newblock \href {https://www.aclweb.org/anthology/N19-1423.pdf} {{BERT}:
  Pre-training of deep bidirectional transformers for language understanding}.
\newblock In \emph{NAACL-HLT}.

\bibitem[{Fang et~al.(2020)Fang, Wang, Gan, Sun, and Liu}]{fang2020filter}
Yuwei Fang, Shuohang Wang, Zhe Gan, Siqi Sun, and Jingjing Liu. 2020.
\newblock \href {https://arxiv.org/pdf/2009.05166.pdf} {{FILTER}: An enhanced
  fusion method for cross-lingual language understanding}.
\newblock \emph{arXiv preprint arXiv:2009.05166}.

\bibitem[{Hu et~al.(2020)Hu, Ruder, Siddhant, Neubig, Firat, and
  Johnson}]{hu2020xtreme}
Junjie Hu, Sebastian Ruder, Aditya Siddhant, Graham Neubig, Orhan Firat, and
  Melvin Johnson. 2020.
\newblock \href {https://arxiv.org/pdf/2003.11080.pdf} {{XTREME}: A massively
  multilingual multi-task benchmark for evaluating cross-lingual
  generalization}.
\newblock \emph{arXiv preprint arXiv:2003.11080}.

\bibitem[{Huang et~al.(2019)Huang, Liang, Duan, Gong, Shou, Jiang, and
  Zhou}]{huang2019Unicoder}
Haoyang Huang, Yaobo Liang, Nan Duan, Ming Gong, Linjun Shou, Daxin Jiang, and
  Ming Zhou. 2019.
\newblock \href {https://arxiv.org/pdf/1909.00964.pdf} {Unicoder: A universal
  language encoder by pre-training with multiple cross-lingual tasks}.
\newblock \emph{arXiv preprint arXiv:1909.00964}.

\bibitem[{Kudo and Richardson(2018)}]{kudo2018sentencepiece}
Taku Kudo and John Richardson. 2018.
\newblock \href {https://arxiv.org/pdf/1808.06226.pdf} {Sentencepiece: A simple
  and language independent subword tokenizer and detokenizer for neural text
  processing}.
\newblock \emph{arXiv preprint arXiv:1808.06226}.

\bibitem[{Lample and Conneau(2019)}]{conneau2019xlm}
Guillaume Lample and Alexis Conneau. 2019.
\newblock \href {https://arxiv.org/pdf/1901.07291.pdf} {Cross-lingual language
  model pretraining}.
\newblock \emph{arXiv preprint arXiv:1901.07291}.

\bibitem[{Lewis et~al.(2019)Lewis, Liu, Goyal, Ghazvininejad, Mohamed, Levy,
  Stoyanov, and Zettlemoyer}]{lewis2019bart}
Mike Lewis, Yinhan Liu, Naman Goyal, Marjan Ghazvininejad, Abdelrahman Mohamed,
  Omer Levy, Ves Stoyanov, and Luke Zettlemoyer. 2019.
\newblock \href {https://arxiv.org/pdf/1910.13461.pdf} {{BART}: Denoising
  sequence-to-sequence pre-training for natural language generation,
  translation, and comprehension}.
\newblock \emph{arXiv preprint arXiv:1910.13461}.

\bibitem[{Lewis et~al.(2020)Lewis, Oğuz, Rinott, Riedel, and
  Schwenk}]{Lewis2020mlqa}
Patrick Lewis, Barlas Oğuz, Ruty Rinott, Sebastian Riedel, and Holger Schwenk.
  2020.
\newblock {MLQA: Evaluating Cross-lingual Extractive Question Answering}.
\newblock In \emph{Proceedings of ACL 2020}.

\bibitem[{Liang et~al.(2020)Liang, Duan, Gong, Wu, Guo, Qi, Gong, Shou, Jiang,
  Cao et~al.}]{liang2020xglue}
Yaobo Liang, Nan Duan, Yeyun Gong, Ning Wu, Fenfei Guo, Weizhen Qi, Ming Gong,
  Linjun Shou, Daxin Jiang, Guihong Cao, et~al. 2020.
\newblock \href {https://arxiv.org/pdf/2004.01401.pdf} {{XGLUE}: A new
  benchmark dataset for cross-lingual pre-training, understanding and
  generation}.
\newblock \emph{arXiv preprint arXiv:2004.01401}.

\bibitem[{Lin et~al.(2020)Lin, Pan, Wang, Qiu, Feng, Zhou, and Li}]{mRASP}
Zehui Lin, Xiao Pan, Mingxuan Wang, Xipeng Qiu, Jiangtao Feng, Hao Zhou, and
  Lei Li. 2020.
\newblock \href {https://doi.org/10.18653/v1/2020.emnlp-main.210} {Pre-training
  multilingual neural machine translation by leveraging alignment information}.
\newblock In \emph{Proceedings of the 2020 Conference on Empirical Methods in
  Natural Language Processing, {EMNLP} 2020, Online, November 16-20, 2020},
  pages 2649--2663. Association for Computational Linguistics.

\bibitem[{Liu et~al.(2020{\natexlab{a}})Liu, Duh, Liu, and
  Gao}]{liu2020deeptransformer}
Xiaodong Liu, Kevin Duh, Liyuan Liu, and Jianfeng Gao. 2020{\natexlab{a}}.
\newblock \href {https://arxiv.org/pdf/2008.07772.pdf} {Very deep transformers
  for neural machine translation}.
\newblock \emph{arXiv preprint arXiv:2008.07772}.

\bibitem[{Liu et~al.(2020{\natexlab{b}})Liu, Gu, Goyal, Li, Edunov,
  Ghazvininejad, Lewis, and Zettlemoyer}]{liu2020mBART}
Yinhan Liu, Jiatao Gu, Naman Goyal, Xian Li, Sergey Edunov, Marjan
  Ghazvininejad, Mike Lewis, and Luke Zettlemoyer. 2020{\natexlab{b}}.
\newblock \href {https://arxiv.org/pdf/1910.13461.pdf} {Multilingual denoising
  pre-training for neural machine translation}.
\newblock \emph{arXiv preprint arXiv:2001.08210}.

\bibitem[{Liu et~al.(2019)Liu, Ott, Goyal, Du, Joshi, Chen, Levy, Lewis,
  Zettlemoyer, and Stoyanov}]{liu2019roberta}
Yinhan Liu, Myle Ott, Naman Goyal, Jingfei Du, Mandar Joshi, Danqi Chen, Omer
  Levy, Mike Lewis, Luke Zettlemoyer, and Veselin Stoyanov. 2019.
\newblock \href {https://arxiv.org/pdf/1907.11692.pdf} {Ro{BERT}a: A robustly
  optimized {BERT} pretraining approach}.
\newblock \emph{arXiv preprint arXiv:1907.11692}.

\bibitem[{Nivre et~al.(2018)Nivre, Abrams, Agi{\'c}, Ahrenberg, Antonsen,
  Aranzabe, Arutie, Asahara, Ateyah, Attia et~al.}]{nivre2018universal}
Joakim Nivre, Mitchell Abrams, {\v{Z}}eljko Agi{\'c}, Lars Ahrenberg, Lene
  Antonsen, Maria~Jesus Aranzabe, Gashaw Arutie, Masayuki Asahara, Luma Ateyah,
  Mohammed Attia, et~al. 2018.
\newblock Universal dependencies 2.2.

\bibitem[{Pan et~al.(2017)Pan, Zhang, May, Nothman, Knight, and Ji}]{Pan2017}
Xiaoman Pan, Boliang Zhang, Jonathan May, Joel Nothman, Kevin Knight, and Heng
  Ji. 2017.
\newblock \href {https://doi.org/10.18653/v1/P17-1178} {{Cross-lingual name
  tagging and linking for 282 languages}}.
\newblock In \emph{Proceedings of ACL 2017}, pages 1946--1958.

\bibitem[{Post(2018)}]{Post18sacreBLEU}
Matt Post. 2018.
\newblock \href {https://arxiv.org/pdf/1804.08771.pdf} {A call for clarity in
  reporting {BLEU} scores}.
\newblock In \emph{Proceedings of the Third Conference on Machine Translation}.

\bibitem[{Ren et~al.(2019)Ren, Wu, Liu, Zhou, and Ma}]{ren19explicit}
Shuo Ren, Yu~Wu, Shujie Liu, Ming Zhou, and Shuai Ma. 2019.
\newblock \href {http://arxiv.org/abs/1909.00180} {Explicit cross-lingual
  pre-training for unsupervised machine translation}.
\newblock volume abs/1909.00180.

\bibitem[{Siddhant et~al.(2020)Siddhant, Johnson, Tsai, Ari, Riesa, Bapna,
  Firat, and Raman}]{siddhant2019mmte}
Aditya Siddhant, Melvin Johnson, Henry Tsai, Naveen Ari, Jason Riesa, Ankur
  Bapna, Orhan Firat, and Karthik Raman. 2020.
\newblock \href {https://aaai.org/ojs/index.php/AAAI/article/view/6414}
  {Evaluating the cross-lingual effectiveness of massively multilingual neural
  machine translation}.
\newblock In \emph{The Thirty-Fourth {AAAI} Conference on Artificial
  Intelligence, {AAAI} 2020}, pages 8854--8861.

\bibitem[{Vaswani et~al.(2017)Vaswani, Shazeer, Parmar, Uszkoreit, Jones,
  Gomez, Kaiser, and Polosukhin}]{vaswani2017attention}
Ashish Vaswani, Noam Shazeer, Niki Parmar, Jakob Uszkoreit, Llion Jones,
  Aidan~N Gomez, {\L}ukasz Kaiser, and Illia Polosukhin. 2017.
\newblock \href
  {https://papers.nips.cc/paper/7181-attention-is-all-you-need.pdf} {Attention
  is all you need}.
\newblock In \emph{Advances in Neural Information Processing Systems}.

\bibitem[{Wenzek et~al.(2019)Wenzek, Lachaux, Conneau, Chaudhary, Guzman,
  Joulin, and Grave}]{wenzek2019ccnet}
Guillaume Wenzek, Marie-Anne Lachaux, Alexis Conneau, Vishrav Chaudhary,
  Francisco Guzman, Armand Joulin, and Edouard Grave. 2019.
\newblock \href {https://arxiv.org/pdf/1911.00359.pdf} {Ccnet: Extracting high
  quality monolingual datasets from web crawl data}.
\newblock \emph{arXiv preprint arXiv:1911.00359}.

\bibitem[{Yang et~al.(2020)Yang, Ma, Zhang, Wu, Li, and Zhou}]{yang2020ALM}
Jian Yang, Shuming Ma, Dongdong Zhang, Shuangzhi Wu, Zhoujun Li, and Ming Zhou.
  2020.
\newblock \href {https://aaai.org/ojs/index.php/AAAI/article/view/6480/6336}
  {Alternating language modeling for cross-lingual pre-training.}
\newblock In \emph{Proceedings of the AAAI Conference on Artificial
  Intelligence}.

\bibitem[{Yang et~al.(2019)Yang, Zhang, Tar, and Baldridge}]{Yang2019paws-x}
Yinfei Yang, Yuan Zhang, Chris Tar, and Jason Baldridge. 2019.
\newblock \href {https://doi.org/10.18653/v1/D19-1382} {{PAWS-X:} {A}
  cross-lingual adversarial dataset for paraphrase identification}.
\newblock In \emph{Proceedings of the 2019 Conference on Empirical Methods in
  Natural Language Processing and the 9th International Joint Conference on
  Natural Language Processing, {EMNLP-IJCNLP} 2019, Hong Kong, China, November
  3-7, 2019}, pages 3685--3690. Association for Computational Linguistics.

\bibitem[{Zhu et~al.(2020)Zhu, Xia, Wu, He, Qin, Zhou, Li, and
  Liu}]{IncorporatingBERT}
Jinhua Zhu, Yingce Xia, Lijun Wu, Di~He, Tao Qin, Wengang Zhou, Houqiang Li,
  and Tie{-}Yan Liu. 2020.
\newblock \href {https://openreview.net/forum?id=Hyl7ygStwB} {Incorporating
  {BERT} into neural machine translation}.
\newblock In \emph{8th International Conference on Learning Representations,
  {ICLR} 2020, Addis Ababa, Ethiopia, April 26-30, 2020}. OpenReview.net.

\bibitem[{Zweigenbaum et~al.(2017)Zweigenbaum, Sharoff, and
  Rapp}]{zweigenbaum2018overview}
Pierre Zweigenbaum, Serge Sharoff, and Reinhard Rapp. 2017.
\newblock \href {https://doi.org/10.18653/v1/w17-2512} {Overview of the second
  {BUCC} shared task: Spotting parallel sentences in comparable corpora}.
\newblock In \emph{Proceedings of the 10th Workshop on Building and Using
  Comparable Corpora, BUCC@ACL 2017, Vancouver, Canada, August 3, 2017}, pages
  60--67. Association for Computational Linguistics.

\end{thebibliography}

\appendix

\section{Pre-Training Details} \label{appendix:pre-training_details}

\begin{table*}[]
\centering
\resizebox{0.5\linewidth}{!}{
\begin{tabular}{c|rrr}
\toprule
\textbf{Language} & \textbf{\#Document(M)} & \textbf{\#Sentence(M)} & \textbf{Size(GB)} \\ \midrule
af & 0.023 & 0.522 & 0.107 \\
ar & 2.823 & 42.659 & 11.786 \\
bg & 0.919 & 14.743 & 5.217 \\
bn & 0.750 & 9.217 & 4.264 \\
cs & 3.980 & 55.754 & 9.668 \\
de & 21.410 & 310.942 & 66.333 \\
el & 1.740 & 24.334 & 9.737 \\
en & 130.087 & 2,215.534 & 479.099 \\
es & 17.569 & 267.764 & 58.774 \\
et & 0.347 & 5.252 & 0.877 \\
eu & 0.342 & 5.216 & 0.613 \\
fr & 15.819 & 267.888 & 58.023 \\
fa & 2.506 & 43.570 & 13.831 \\
fi & 1.530 & 23.790 & 3.940 \\
fy & 0.027 & 0.537 & 0.054 \\
gu & 0.039 & 0.519 & 0.228 \\
gd & 0.009 & 0.126 & 0.020 \\
he & 0.755 & 12.338 & 3.073 \\
hi & 0.536 & 7.303 & 3.762 \\
hu & 1.816 & 29.962 & 6.421 \\
id & 3.417 & 60.908 & 11.528 \\
it & 9.336 & 133.006 & 30.854 \\
ja & 27.967 & 588.926 & 71.785 \\
jv & 0.002 & 0.138 & 0.030 \\
ka & 0.141 & 1.756 & 0.766 \\
kk & 0.061 & 1.545 & 0.448 \\
ko & 11.609 & 227.396 & 27.837 \\
lt & 0.552 & 7.996 & 1.480 \\
lv & 0.281 & 4.159 & 0.798 \\
ms & 0.334 & 3.762 & 0.455 \\
ml & 0.162 & 2.615 & 1.025 \\
my & 0.045 & 0.893 & 0.306 \\
mr & 0.059 & 0.708 & 0.365 \\
pl & 6.642 & 93.760 & 19.082 \\
pt & 8.623 & 128.107 & 25.612 \\
ne & 0.080 & 0.829 & 0.429 \\
nl & 6.513 & 85.997 & 16.648 \\
ru & 35.887 & 580.291 & 203.105 \\
ro & 1.944 & 31.929 & 7.056 \\
si & 0.132 & 2.927 & 0.902 \\
sw & 0.057 & 0.945 & 0.179 \\
ta & 0.876 & 20.376 & 6.422 \\
te & 0.288 & 4.995 & 1.721 \\
tr & 18.547 & 291.081 & 40.321 \\
th & 6.278 & 117.826 & 27.941 \\
tl & 0.166 & 5.611 & 0.679 \\
vi & 12.183 & 234.071 & 37.919 \\
ur & 0.460 & 7.509 & 2.003 \\
yo & 0.0002 & 0.003 & 0.0005 \\
zh & 27.067 & 497.408 & 87.005 \\ \midrule
\textbf{Total} & 382.735 & 6,475.444 & 1,360.526 \\
\bottomrule
\end{tabular}
}
\caption{The statistics of monolingual pre-training corpus.}
\label{tb:mono-stat}
\end{table*}
\begin{table*}[]
\centering
\resizebox{0.98\linewidth}{!}{
\begin{tabular}{cr|cr|cr|cr|cr|cr|cr|cr|cr}
\toprule
\textbf{Pair} & \textbf{\#Sent(K)} & \textbf{Pair} & \textbf{\#Sent(K)} & \textbf{Pair} & \textbf{\#Sent(K)} & \textbf{Pair} & \textbf{\#Sent(K)} & \textbf{Pair} & \textbf{\#Sent(K)} & \textbf{Pair} & \textbf{\#Sent(K)} & \textbf{Pair} & \textbf{\#Sent(K)} & \textbf{Pair} & \textbf{\#Sent(K)} & \textbf{Pair} & \textbf{\#Sent(K)} \\ \midrule 
af-ar & 12.34 & bg-my & 0.08 & de-he & 12751.69 & en-tr & 46584.82 & eu-zh & 19.76 & fy-vi & 34.95 & id-pt & 6825.29 & ko-sw & 6.74 & pl-es & 46863.47\\ 
af-bg & 18.19 & bg-ne & 0.01 & de-hi & 106.11 & en-ur & 781.60 & fa-fi & 4485.62 & gd-es & 21.62 & id-ro & 7944.59 & ko-ta & 13.74 & pl-pt & 72437.93\\ 
af-bn & 1.19 & bg-nl & 30757.50 & de-hu & 24409.40 & en-vi & 3563.39 & fa-fr & 4507.06 & gd-it & 13.26 & id-ru & 5039.44 & ko-te & 0.93 & pl-ru & 19170.23\\ 
af-cs & 17.93 & bg-pl & 33043.03 & de-id & 4786.89 & en-yo & 0.13 & fa-he & 4944.80 & gd-pl & 12.29 & id-si & 366.00 & ko-th & 230.84 & pl-sw & 1424.02\\ 
af-de & 19.28 & bg-pt & 30058.54 & de-it & 35936.62 & en-zh & 28952.02 & fa-hi & 186.23 & gd-pt & 18.90 & id-sw & 30.56 & ko-tl & 1.21 & pl-tl & 1039.37\\ 
af-el & 29.83 & bg-ro & 38925.52 & de-ja & 1472.72 & es-et & 18090.74 & fa-hu & 5201.51 & gd-ru & 10.39 & id-ta & 35.37 & ko-tr & 1246.58 & pl-tr & 32470.18\\ 
af-en & 44.70 & bg-ru & 17423.43 & de-ka & 123.12 & es-eu & 793.59 & fa-id & 3220.00 & gd-tr & 14.12 & id-te & 13.30 & ko-ur & 57.21 & pl-ur & 391.99\\ 
af-es & 34.31 & bg-si & 460.50 & de-kk & 3.72 & es-fa & 5696.70 & fa-it & 4243.56 & he-hi & 57.85 & id-th & 1562.94 & ko-vi & 345.79 & pl-vi & 3790.71\\ 
af-et & 6.34 & bg-sw & 10.80 & de-ko & 776.89 & es-fi & 34222.07 & fa-ja & 1072.14 & he-hu & 23959.87 & id-tl & 7.80 & ko-zh & 56.43 & pt-ro & 33802.95\\ 
af-fa & 3.07 & bg-ta & 27.14 & de-lt & 9134.99 & es-fr & 96233.21 & fa-ka & 96.32 & he-id & 6362.29 & id-tr & 8017.99 & lt-lv & 6546.76 & pt-ru & 14698.48\\ 
af-fi & 10.25 & bg-te & 17.14 & de-lv & 8532.06 & es-he & 27060.49 & fa-kk & 1.01 & he-it & 19908.66 & id-ur & 172.71 & lt-ml & 66.40 & pt-si & 450.40\\ 
af-fr & 18.56 & bg-th & 2733.84 & de-ml & 294.16 & es-hi & 85.35 & fa-ko & 627.97 & he-ja & 1683.29 & id-vi & 2081.70 & lt-ms & 393.89 & pt-sw & 13.06\\ 
af-fy & 36.94 & bg-tl & 6.69 & de-ms & 1228.82 & es-hu & 43947.78 & fa-lt & 615.78 & he-ka & 149.06 & id-zh & 356.46 & lt-nl & 7497.18 & pt-ta & 26.37\\ 
af-he & 14.53 & bg-tr & 31179.35 & de-my & 0.68 & es-id & 8015.69 & fa-lv & 228.40 & he-kk & 2.38 & it-ja & 1613.05 & lt-pl & 9965.36 & pt-te & 19.32\\ 
af-hi & 1.15 & bg-ur & 71.60 & de-ne & 0.28 & es-it & 49423.51 & fa-ml & 308.49 & he-ko & 1094.72 & it-ka & 106.70 & lt-pt & 7663.84 & pt-th & 2561.09\\ 
af-hu & 16.32 & bg-vi & 2855.13 & de-nl & 34909.49 & es-ja & 1929.41 & fa-ms & 1072.22 & he-lt & 1220.91 & it-kk & 2.54 & lt-ro & 5786.22 & pt-tl & 10.35\\ 
af-id & 4.56 & bg-zh & 746.27 & de-pt & 32610.10 & es-ka & 181.19 & fa-my & 0.06 & he-lv & 461.81 & it-ko & 1125.97 & lt-ru & 950.02 & pt-tr & 27428.79\\ 
af-it & 15.01 & bn-cs & 340.51 & de-ro & 24261.82 & es-kk & 2.48 & fa-ne & 0.01 & he-ml & 250.07 & it-lt & 7359.92 & lt-si & 106.53 & pt-ur & 73.57\\ 
af-ja & 1.98 & bn-de & 346.51 & de-ru & 10904.25 & es-ko & 1229.50 & fa-nl & 5010.64 & he-ms & 1455.61 & it-lv & 6607.27 & lt-sw & 0.02 & pt-vi & 2963.83\\ 
af-lt & 0.65 & bn-el & 340.94 & de-si & 324.86 & es-lt & 7702.99 & fa-pt & 4998.09 & he-my & 0.05 & it-ml & 235.96 & lt-ta & 13.04 & pt-yo & 0.05\\ 
af-lv & 1.08 & bn-en & 752.08 & de-sw & 45.61 & es-lv & 6703.10 & fa-ro & 5714.73 & he-nl & 22186.61 & it-ms & 1269.97 & lt-te & 9.71 & pt-zh & 846.44\\ 
af-ml & 2.18 & bn-es & 480.35 & de-ta & 42.32 & es-ml & 339.71 & fa-ru & 4205.20 & he-pl & 24962.23 & it-my & 0.36 & lt-th & 263.89 & ro-ru & 19568.56\\ 
af-ms & 1.31 & bn-et & 252.68 & de-te & 12.81 & es-ms & 1731.36 & fa-si & 292.78 & he-pt & 21226.36 & it-ne & 1.02 & lt-tl & 1.36 & ro-si & 504.24\\ 
af-nl & 22.61 & bn-eu & 42.42 & de-th & 1695.53 & es-my & 2.50 & fa-sw & 69.51 & he-ro & 26370.15 & it-nl & 37644.29 & lt-tr & 1377.40 & ro-sw & 10.72\\ 
af-pl & 1096.89 & bn-fa & 391.89 & de-tl & 12.91 & es-ne & 2.87 & fa-ta & 83.30 & he-ru & 14873.77 & it-pl & 35037.31 & lt-ur & 4.47 & ro-ta & 33.50\\ 
af-pt & 22.68 & bn-fi & 279.35 & de-tr & 17579.53 & es-nl & 46908.79 & fa-te & 10.11 & he-si & 435.87 & it-pt & 35301.98 & lt-vi & 486.84 & ro-te & 24.44\\ 
af-ro & 32.19 & bn-fr & 373.13 & de-ur & 218.89 & es-pt & 47542.26 & fa-th & 1201.04 & he-sw & 0.06 & it-ro & 32153.38 & lt-zh & 40.65 & ro-th & 2874.73\\ 
af-ru & 15.41 & bn-he & 302.62 & de-vi & 2284.70 & es-ro & 48229.60 & fa-tl & 7.02 & he-ta & 23.99 & it-ru & 17669.12 & lv-ml & 23.32 & ro-tl & 8.61\\ 
af-si & 0.98 & bn-hi & 38.68 & de-zh & 587.96 & es-ru & 55569.05 & fa-tr & 6217.24 & he-te & 18.65 & it-si & 366.97 & lv-ms & 163.28 & ro-tr & 36549.61\\ 
af-ta & 1.13 & bn-hu & 321.36 & el-en & 55078.46 & es-si & 512.22 & fa-ur & 568.00 & he-th & 2666.00 & it-sw & 15.77 & lv-nl & 6622.81 & ro-ur & 73.55\\ 
af-th & 2.08 & bn-id & 360.65 & el-es & 46876.21 & es-sw & 41.33 & fa-vi & 1514.04 & he-tl & 6.58 & it-ta & 17.39 & lv-pl & 9460.93 & ro-vi & 3207.73\\ 
af-tr & 24.22 & bn-it & 301.31 & el-et & 16463.57 & es-ta & 31.19 & fa-zh & 372.10 & he-tr & 25179.32 & it-te & 9.93 & lv-pt & 6672.14 & ro-zh & 947.91\\ 
af-vi & 3.30 & bn-ja & 142.19 & el-eu & 673.93 & es-te & 21.76 & fi-fr & 28973.81 & he-ur & 20.57 & it-th & 2447.55 & lv-ro & 4833.77 & ru-si & 340.11\\ 
ar-bg & 23090.32 & bn-ka & 8.68 & el-fa & 5137.52 & es-th & 2976.49 & fi-he & 17820.49 & he-vi & 2813.73 & it-tl & 13.30 & lv-ru & 435.73 & ru-sw & 84.77\\ 
ar-bn & 378.28 & bn-ko & 93.92 & el-fi & 28885.65 & es-tl & 13.55 & fi-hi & 55.60 & he-zh & 563.24 & it-tr & 25770.29 & lv-si & 34.42 & ru-ta & 61.50\\ 
ar-cs & 24147.25 & bn-lt & 96.24 & el-fr & 38560.84 & es-tr & 39805.02 & fi-hu & 27350.30 & hi-hu & 60.05 & it-ur & 69.89 & lv-sw & 0.01 & ru-te & 10.80\\ 
ar-de & 12733.65 & bn-lv & 41.21 & el-he & 22042.85 & es-ur & 79.44 & fi-id & 5806.36 & hi-id & 85.85 & it-vi & 2542.41 & lv-ta & 4.10 & ru-th & 2194.91\\ 
ar-el & 22486.60 & bn-ml & 93.14 & el-hi & 62.26 & es-vi & 3215.16 & fi-it & 26756.85 & hi-it & 60.12 & it-yo & 0.10 & lv-te & 4.01 & ru-tl & 13.43\\ 
ar-en & 60392.55 & bn-ms & 203.84 & el-hu & 34559.75 & es-yo & 0.12 & fi-ja & 1599.82 & hi-ja & 46.14 & it-zh & 473.74 & lv-th & 108.92 & ru-tr & 19317.60\\ 
ar-es & 57561.29 & bn-my & 0.78 & el-id & 7098.25 & es-zh & 28688.60 & fi-ka & 148.42 & hi-ka & 0.80 & ja-ka & 35.37 & lv-tr & 515.30 & ru-ur & 417.23\\ 
ar-et & 9738.71 & bn-ne & 0.78 & el-it & 34337.63 & et-eu & 406.33 & fi-kk & 3.41 & hi-ko & 33.66 & ja-kk & 1.21 & lv-ur & 1.08 & ru-vi & 2289.72\\ 
ar-eu & 578.30 & bn-nl & 331.34 & el-ja & 1740.08 & et-fa & 3085.41 & fi-ko & 859.31 & hi-lt & 23.67 & ja-ko & 308.30 & lv-vi & 209.40 & ru-yo & 0.10\\ 
ar-fa & 5679.85 & bn-pt & 333.59 & el-ka & 167.39 & et-fi & 15969.08 & fi-lt & 7507.00 & hi-lv & 12.61 & ja-lt & 281.74 & lv-zh & 14.71 & ru-zh & 28138.59\\ 
ar-fi & 17169.90 & bn-ro & 337.94 & el-kk & 2.33 & et-fr & 15697.59 & fi-lv & 6732.38 & hi-ml & 30.28 & ja-lv & 99.97 & ml-ms & 101.75 & si-ta & 6.33\\ 
ar-fr & 50632.52 & bn-ru & 392.15 & el-ko & 1130.94 & et-fy & 51.63 & fi-ml & 232.48 & hi-ms & 40.38 & ja-ml & 79.78 & ml-nl & 268.10 & si-te & 1.85\\ 
ar-he & 20577.16 & bn-si & 47.49 & el-lt & 7400.42 & et-he & 9814.49 & fi-ms & 1276.96 & hi-my & 0.01 & ja-ms & 489.33 & ml-pt & 280.62 & si-th & 109.38\\ 
ar-hi & 96.26 & bn-sw & 23.91 & el-lv & 6549.40 & et-hi & 43.98 & fi-nl & 30693.72 & hi-ne & 0.04 & ja-nl & 1716.42 & ml-ro & 325.97 & si-tl & 3.02\\ 
ar-hu & 23770.38 & bn-ta & 15.67 & el-ml & 302.85 & et-hu & 16819.43 & fi-pl & 29451.87 & hi-nl & 92.46 & ja-pl & 3295.60 & ml-ru & 310.59 & si-tr & 492.12\\ 
ar-id & 6989.56 & bn-th & 129.60 & el-ms & 1547.63 & et-id & 4282.23 & fi-pt & 29269.50 & hi-pl & 681.08 & ja-pt & 1756.87 & ml-si & 28.01 & si-ur & 4.95\\ 
ar-it & 20070.27 & bn-tl & 2.05 & el-my & 0.55 & et-it & 14462.11 & fi-ro & 27988.13 & hi-pt & 62.44 & ja-ro & 1843.14 & ml-sw & 12.47 & si-vi & 210.15\\ 
ar-ja & 1847.98 & bn-tr & 441.74 & el-ne & 1.04 & et-ja & 1176.51 & fi-ru & 12403.26 & hi-ro & 82.89 & ja-ru & 1491.65 & ml-ta & 15.90 & si-zh & 14.28\\ 
ar-ka & 161.65 & bn-ur & 108.74 & el-nl & 37188.78 & et-ka & 110.02 & fi-si & 391.99 & hi-ru & 142.53 & ja-si & 162.96 & ml-th & 81.03 & sw-ta & 6.24\\ 
ar-kk & 1.28 & bn-vi & 219.57 & el-pt & 35491.54 & et-kk & 1.14 & fi-sw & 0.02 & hi-si & 11.41 & ja-sw & 6.24 & ml-tl & 3.30 & sw-th & 6.24\\ 
ar-ko & 1262.60 & bn-zh & 85.24 & el-ro & 37986.26 & et-ko & 492.79 & fi-ta & 20.08 & hi-sw & 12.52 & ja-ta & 18.92 & ml-tr & 439.25 & sw-tr & 91.95\\ 
ar-lt & 1177.67 & cs-de & 24049.84 & el-ru & 17052.36 & et-lt & 7431.17 & fi-te & 17.13 & hi-ta & 41.00 & ja-te & 5.68 & ml-ur & 100.52 & sw-ur & 50.29\\ 
ar-lv & 433.66 & cs-el & 35372.28 & el-si & 466.44 & et-lv & 6728.85 & fi-th & 2288.65 & hi-te & 23.18 & ja-th & 632.26 & ml-vi & 124.30 & sw-yo & 0.03\\ 
ar-ml & 348.33 & cs-en & 54470.47 & el-sw & 4.85 & et-ml & 179.99 & fi-tl & 5.91 & hi-th & 37.53 & ja-tl & 10.06 & ml-zh & 34.77 & sw-zh & 19.31\\ 
ar-ms & 1555.33 & cs-es & 44962.42 & el-ta & 20.44 & et-ms & 1135.84 & fi-tr & 22551.99 & hi-tl & 0.51 & ja-tr & 1896.56 & ms-nl & 1409.07 & ta-te & 21.16\\ 
ar-my & 0.18 & cs-et & 17819.46 & el-te & 18.10 & et-nl & 16560.63 & fi-ur & 19.43 & hi-tr & 176.39 & ja-ur & 61.41 & ms-pt & 1523.57 & ta-th & 14.15\\ 
ar-ne & 0.41 & cs-eu & 686.53 & el-th & 2505.71 & et-pl & 19633.08 & fi-vi & 2517.08 & hi-ur & 101.10 & ja-vi & 679.31 & ms-ro & 1732.68 & ta-tr & 77.76\\ 
ar-nl & 21273.78 & cs-fa & 5417.48 & el-tl & 10.13 & et-pt & 16768.45 & fi-zh & 630.12 & hi-vi & 32.99 & ja-zh & 104.37 & ms-ru & 1210.56 & ta-ur & 49.89\\ 
ar-pl & 24819.83 & cs-fi & 28031.47 & el-tr & 31048.88 & et-ro & 15880.62 & fr-he & 21218.88 & hi-zh & 25.57 & ka-ko & 17.13 & ms-si & 204.06 & ta-vi & 12.65\\ 
ar-pt & 20379.56 & cs-fr & 34876.02 & el-ur & 24.36 & et-ru & 6630.25 & fr-hi & 68.31 & hu-id & 7253.81 & ka-lt & 30.49 & ms-sw & 8.99 & ta-zh & 13.02\\ 
ar-ro & 26187.15 & cs-he & 24503.29 & el-vi & 2966.14 & et-si & 331.22 & fr-hu & 37027.57 & hu-it & 33513.06 & ka-lv & 10.71 & ms-ta & 15.24 & te-th & 0.96\\ 
ar-ru & 45992.72 & cs-hi & 86.86 & el-yo & 0.11 & et-sw & 0.01 & fr-id & 6235.29 & hu-ja & 1767.63 & ka-ml & 6.56 & ms-te & 4.70 & te-tr & 18.84\\ 
ar-si & 483.96 & cs-hu & 39272.92 & el-zh & 649.81 & et-ta & 14.34 & fr-it & 41162.37 & hu-ka & 165.84 & ka-ms & 31.86 & ms-th & 413.17 & te-vi & 9.34\\ 
ar-sw & 16.52 & cs-id & 7310.27 & en-es & 156560.00 & et-te & 14.44 & fr-ja & 1608.52 & hu-kk & 2.58 & ka-nl & 155.10 & ms-tl & 7.26 & th-tl & 7.28\\ 
ar-ta & 37.15 & cs-it & 33935.96 & en-et & 22284.30 & et-th & 1746.50 & fr-ka & 139.63 & hu-ko & 1168.66 & ka-pt & 165.00 & ms-tr & 1754.22 & th-tr & 3054.07\\ 
ar-te & 19.33 & cs-ja & 1806.97 & en-eu & 805.78 & et-tl & 3.09 & fr-kk & 1.34 & hu-lt & 7623.58 & ka-ro & 182.79 & ms-ur & 68.94 & th-ur & 58.65\\ 
ar-th & 2959.96 & cs-ka & 163.35 & en-fa & 7462.52 & et-tr & 11408.82 & fr-ko & 991.60 & hu-lv & 6776.32 & ka-ru & 104.82 & ms-vi & 851.69 & th-vi & 672.82\\ 
ar-tl & 7.58 & cs-kk & 1.26 & en-fi & 42783.36 & et-ur & 19.52 & fr-lt & 9440.34 & hu-ml & 279.13 & ka-si & 7.96 & ms-zh & 85.86 & th-zh & 133.45\\ 
ar-tr & 26683.62 & cs-ko & 1199.62 & en-fr & 161519.91 & et-vi & 2048.37 & fr-lv & 8569.67 & hu-ms & 1581.43 & ka-th & 43.37 & my-nl & 0.10 & tl-tr & 14.51\\ 
ar-ur & 126.33 & cs-lt & 7694.12 & en-fy & 126.19 & et-zh & 405.30 & fr-ml & 278.47 & hu-my & 0.06 & ka-tl & 1.27 & my-pt & 0.10 & tl-vi & 5.86\\ 
ar-vi & 2875.00 & cs-lv & 6745.84 & en-gd & 47.02 & eu-fa & 245.78 & fr-ms & 1423.08 & hu-nl & 33904.34 & ka-tr & 178.79 & my-ro & 0.03 & tr-ur & 473.08\\ 
ar-yo & 0.01 & cs-ml & 319.93 & en-he & 30028.28 & eu-fi & 581.61 & fr-my & 1.47 & hu-pl & 39869.14 & ka-ur & 1.98 & my-ru & 0.81 & tr-vi & 3178.03\\ 
ar-zh & 28120.22 & cs-ms & 1592.17 & en-hi & 1844.38 & eu-fr & 636.16 & fr-ne & 1.45 & hu-pt & 31715.19 & ka-vi & 53.58 & my-sw & 0.15 & tr-zh & 1029.21\\ 
bg-bn & 310.12 & cs-my & 0.08 & en-hu & 55233.87 & eu-he & 566.71 & fr-nl & 47363.70 & hu-ro & 38807.61 & ka-zh & 6.52 & my-tr & 0.03 & ur-vi & 12.52\\ 
bg-cs & 34502.46 & cs-ne & 0.07 & en-id & 9677.33 & eu-hi & 9.98 & fr-pt & 42850.13 & hu-ru & 19172.99 & kk-lt & 0.83 & my-ur & 0.02 & ur-zh & 99.78\\ 
bg-de & 19852.81 & cs-nl & 34427.07 & en-it & 76257.21 & eu-hu & 663.68 & fr-ro & 37249.80 & hu-si & 460.99 & kk-lv & 1.13 & my-zh & 0.13 & vi-zh & 148.22\\ 
bg-el & 32130.86 & cs-pt & 32469.01 & en-ja & 2177.89 & eu-id & 307.85 & fr-ru & 54231.81 & hu-sw & 0.68 & kk-ms & 1.12 & ne-nl & 0.09\\ 
bg-en & 47247.04 & cs-ro & 39226.31 & en-ka & 199.98 & eu-it & 568.66 & fr-si & 393.48 & hu-ta & 20.63 & kk-nl & 1.85 & ne-pt & 0.38\\ 
bg-es & 39728.55 & cs-ru & 19703.43 & en-kk & 3.71 & eu-ja & 139.14 & fr-sw & 29.32 & hu-te & 17.57 & kk-pl & 77.88 & ne-ro & 0.04\\ 
bg-et & 15188.54 & cs-si & 454.26 & en-ko & 1493.95 & eu-ka & 9.42 & fr-ta & 24.03 & hu-th & 2867.23 & kk-pt & 3.35 & ne-ru & 1.30\\ 
bg-eu & 605.10 & cs-sw & 17.34 & en-lt & 10992.89 & eu-ko & 72.17 & fr-te & 11.93 & hu-tl & 10.79 & kk-ro & 2.35 & ne-sw & 0.05\\ 
bg-fa & 4927.53 & cs-ta & 32.81 & en-lv & 9883.08 & eu-lt & 108.12 & fr-th & 2325.22 & hu-tr & 32494.90 & kk-ru & 2.22 & ne-tr & 0.03\\ 
bg-fi & 25191.01 & cs-te & 18.72 & en-ml & 573.95 & eu-lv & 36.81 & fr-tl & 13.18 & hu-ur & 23.32 & kk-th & 0.93 & ne-ur & 0.06\\ 
bg-fr & 30185.98 & cs-th & 2858.53 & en-ms & 2050.83 & eu-ml & 42.72 & fr-tr & 29245.91 & hu-vi & 2974.61 & kk-tr & 2.59 & ne-zh & 0.01\\ 
bg-he & 22887.40 & cs-tl & 7.44 & en-my & 2.43 & eu-ms & 129.20 & fr-ur & 73.99 & hu-zh & 730.70 & kk-vi & 1.18 & nl-pt & 37775.73\\ 
bg-hi & 71.38 & cs-tr & 32797.28 & en-ne & 2.89 & eu-nl & 619.88 & fr-vi & 2752.32 & id-it & 5831.16 & ko-lt & 148.54 & nl-ro & 36051.60\\ 
bg-hu & 34293.44 & cs-ur & 122.87 & en-nl & 65918.54 & eu-pt & 641.30 & fr-yo & 0.12 & id-ja & 1271.31 & ko-lv & 57.10 & nl-ru & 16582.78\\ 
bg-id & 7047.21 & cs-vi & 3040.14 & en-pl & 59729.77 & eu-ro & 715.99 & fr-zh & 28008.77 & id-ka & 85.07 & ko-ml & 42.92 & nl-si & 410.92\\ 
bg-it & 27649.85 & cs-zh & 894.87 & en-pt & 61861.36 & eu-ru & 435.12 & fy-es & 49.12 & id-kk & 1.03 & ko-ms & 291.25 & nl-sw & 31.38\\ 
bg-ja & 1658.40 & de-el & 30170.64 & en-ro & 60415.46 & eu-si & 34.56 & fy-he & 44.06 & id-ko & 605.78 & ko-my & 0.12 & nl-ta & 39.21\\ 
bg-ka & 193.27 & de-en & 83872.47 & en-ru & 65105.13 & eu-ta & 3.35 & fy-it & 47.88 & id-lt & 855.43 & ko-ne & 0.01 & nl-te & 16.07\\ 
bg-kk & 3.40 & de-es & 41634.80 & en-si & 601.16 & eu-te & 0.73 & fy-ja & 37.61 & id-lv & 342.36 & ko-nl & 1120.75 & nl-th & 2548.14\\ 
bg-ko & 1056.96 & de-et & 15186.40 & en-sw & 171.65 & eu-th & 80.75 & fy-pl & 49.37 & id-ml & 230.67 & ko-pl & 2722.47 & nl-tl & 8.18\\ 
bg-lt & 5604.11 & de-eu & 534.93 & en-ta & 125.96 & eu-tl & 2.60 & fy-pt & 95.81 & id-ms & 1614.63 & ko-pt & 1119.49 & nl-tr & 28822.22\\ 
bg-lv & 4748.15 & de-fa & 3948.14 & en-te & 27.22 & eu-tr & 722.77 & fy-ru & 45.83 & id-my & 0.11 & ko-ro & 1242.76 & nl-ur & 171.71\\ 
bg-ml & 283.77 & de-fi & 25753.06 & en-th & 3375.07 & eu-ur & 2.01 & fy-sw & 0.37 & id-ne & 0.07 & ko-ru & 959.46 & nl-vi & 2748.28\\ 
bg-ms & 1506.56 & de-fr & 44392.06 & en-tl & 16.03 & eu-vi & 201.28 & fy-tr & 45.40 & id-nl & 6493.33 & ko-si & 58.66 & nl-zh & 866.75 & \textbf{Total}	& 6,421,152.04\\ 

\bottomrule
\end{tabular}
}
\caption{The statistics of bilingual (parallel) pre-training corpus.}
\label{tb:bili-stat}
\end{table*}

For monolingual data, following XLM-R~\citep{conneau2019xlmr}, we build a clean CommonCrawl Corpus using an open-source tool CCNet~\citep{wenzek2019ccnet}. There are 1.36TB monolingual data in 50 languages before up/down-sampling.
Table~\ref{tb:mono-stat} reports the language codes and statistics of pre-training data.
We collect bilingual corpus in 50 languages from the OPUS website\footnote{\url{http://opus.nlpl.eu/}}, including MultiUN, UNPC, Bombay, EU-bookshop, OpenSubtitles2018, Tanzil, GlobalVoices, ParaCrawl, MultiParaCrawl, DGT, Tilde, Europarl, Wikipedia, ECB, TED2013, News-Commentary, Ubuntu, Books, UN, infopankki-v1, EUconst, and Bianet. In total, there are 1TB bilingual training data before pre-processing, covering 879 language pairs. Table~\ref{tb:bili-stat} lists the statistics for each language pair.
We then apply subword tokenization directly on raw text data using Sentence Piece Model~\citep{kudo2018sentencepiece} without any additional preprocessing.

We use the whole corpus to train \veco and a subset ($\sim 1/4$) that contains 33 languages to train small-sized XLM, mBART and \veco. The full set of pre-training hyperparameters for small-sized and large-sized \veco (default) are listed in Table~\ref{tab:hyperpre}.

\begin{table*}[!h]
\footnotesize
\centering
\resizebox{0.6\textwidth}{!}{
\begin{tabular}{lcccc}
\toprule
\textbf{Pre-training Hyperparameters} & \textbf{Large} & \textbf{Small} \\
\midrule
Number of layers & 24 & 6 \\
Hidden Size & 1024 & 768\\
FFN inner hidden size & 4096 & 3072\\
Attention heads & 16 & 12\\
Attention head size & 64 & 64\\
Embedding Size & 1024 & 768 \\
Mask percent (monolingual/ bilingual) & 15\%/25\% &  15\%/25\%\\
Learning Rate Decay & Linear & Linear \\
Warmup steps & 12k & 12k \\
Learning Rate & 2e-4 & 3e-4\\
Adam $\epsilon$ & 1e-6 & 1e-6\\
Adam $\beta_1$ & 0.9 & 0.9\\
Adam $\beta_2$ & 0.98 &0.999 \\
Attention Dropout & 0.1 & 0.1\\
Dropout & 0.1 &0.1 \\
Weight Decay & 0.01 & 0.01\\
Max Sequence Length  (monolingual/bilingual)& 512/128 & 512/128\\
Batch Size  (monolingual/bilingual)& 1024/4096 & 1024/4096\\
Train Steps  & 240k & 240k \\
Total Parameters  & 662M & 247M \\
\bottomrule
\end{tabular}
}
\caption{The pre-training hyperparameters.}
\label{tab:hyperpre}
\end{table*}

\section{More details about Illustrated Attention}
The models illustrated with attention patterns in Figure 1 of main paper (not appendix), are the base-sized XLM~\footnote{\url{https://huggingface.co/xlm-mlm-tlm-xnli15-1024}} and XLM-R~\footnote{\url{https://huggingface.co/xlm-roberta-base}}. We show the attention scores averaged on all heads in the middle layer.

\section{Fine-Tuning Details on XTERME}
We select the model with the best average result over all the languages on the dev sets, by searching the learning rate over [5e-6,8e-6,1e-5,2e-5,3e-5] for the \textit{Cross-lingual Transfer} setting and [5e-6,6e-6,7e-6,8e-6,9e-6] for \textit{Translate-Train-All} setting, training epoch over [3,5,10], and batch size over [16,32,64].

\section{Detailed Results on XTREME}
\label{app:detailed_results}
The detailed results of each XTREME task under the cross-lingual transfer and translate-train-all settings on all languages are listed in the following tables.
\begin{table*}[!h]
\resizebox{\textwidth}{!}{
\begin{tabular}{lcccccccccccccccc}
\toprule
\textbf{Model}  & en   & ar   & bg   & de   & el   & es   & fr   & hi   & ru   & sw   & th   & tr   & ur   & vi   & zh   & \textbf{Avg.}  \\
\midrule
\multicolumn{17}{l}{\textit{\textbf{Cross-lingual Transfer}}}\\
XLM-R  & 88.7 & 77.2 & 83.0 & 82.5 & 80.8 & 83.7 & 82.2 & 75.6 & 79.1 & 71.2 & 77.4 & 78.0 & 71.7 & 79.3 & 78.2 & 79.2 \\
\vecoplugout & 88.2 & 79.2 & 83.1 & 82.9 & 81.2 & 84.2 & 82.8 & 76.2 & 80.3 & 74.3 & 77.0 & 78.4 & 71.3 & 80.4 & 79.1 & 79.9 \\
\midrule
\multicolumn{17}{l}{\textit{\textbf{Translate-Train-All}}} \\
XLM-R & 88.6 & 82.2 & 85.2 & 84.5 & 84.5 & 85.7 & 84.2 & 80.8 & 81.8 & 77.0 & 80.2 & 82.1 & 77.7 & 82.6 & 82.7 & 82.6 \\   
\vecoplugout & 88.9 & 82.4 & 86.0 & 84.7 & 85.3 & 86.2 & 85.8 & 80.1 & 83.0 & 77.2 & 80.9 & 82.8 & 75.3 & 83.1 & 83.0 & 83.0 \\
\vecoplugin & 89.3 & 83.7 & 87.0 & 85.9 & 85.8 & 87.3 & 86.7 & 81.8 & 83.6 & 79.9 & 82.5 & 84.3 & 77.7 & 84.4 & 84.0 & 84.3 \\
\bottomrule
\end{tabular}}
\caption{XNLI accuracy scores for each language. }
\label{tab:xnli_results}
\end{table*}

\begin{table*}[!h]
\begin{minipage}[]{0.61\textwidth}
\centering
\small
\vspace{-0.1in}
\resizebox{\columnwidth}{!}{
\begin{tabular}{lcccccccc}
\toprule
Model & en & de & es & fr & ja & ko & zh & \textbf{Avg.} \\
\midrule
\multicolumn{9}{l}{\textit{\textbf{Cross-lingual Transfer}}}\\
XLM-R & 94.7& 89.7& 90.1& 90.4& 78.7& 79.0& 82.3 & 86.4 \\
\vecoplugout & 96.2 & 91.3 & 91.4 & 92.0 & 81.8 & 82.9 & 85.1 & 88.7 \\
\midrule
\multicolumn{9}{l}{\textit{\textbf{Translate-Train-All}}} \\
\vecoplugout & 96.4 & 93.0 & 93.0 & 93.5 & 87.2 & 86.8 & 87.9 & 91.1 \\
\vecoplugin & 96.5 & 94.4 & 94.3 & 94.0 & 89.0 & 90.3 & 91.0 & 92.8 \\ 
\bottomrule
\end{tabular}}
\caption{PAWS-X accuracy scores.}
\label{tbl:PAWS}
\end{minipage}
\hfill
\begin{minipage}[]{0.39\textwidth}
\centering
\label{tab:bucc_results}
\resizebox{\columnwidth}{!}{
\begin{tabular}{lccccc}
\toprule
Model & de   & fr   & ru   & zh   & \textbf{Avg.}  \\ 
\midrule
\multicolumn{6}{l}{\textit{\textbf{Cross-lingual Transfer}}}\\
XLM-R  & 67.5  & 66.5 & 73.5 & 56.7 & 66.0 \\
\vecoplugout & 89.6 & 84.6 & 87.4 & 78.5 & 85.0 \\
\midrule
\multicolumn{6}{l}{\textit{\textbf{Translate-Train-All}}} \\
\vecoplugout & 93.0 & 88.7 & 89.9 & 85.7 & 89.3 \\
\vecoplugin & 95.4 & 91.9 & 93.1 & 89.9 & 92.6 \\
\bottomrule
\end{tabular}}
\caption{BUCC F1 results.}
\end{minipage}
\hfill
\end{table*}

\begin{table*}[!h]
\centering
\resizebox{\linewidth}{!}{
\begin{tabular}{lccccccccccccccccc}
\toprule
Model & af & ar & bg & de & el & en & es & et & eu & fa &  fi & fr & he & hi & hu & id & it \\
\midrule
\multicolumn{17}{l}{\textit{\textbf{Cross-lingual Transfer}}}\\
XLM-R & 89.8 & 67.5 & 88.1 & 88.5& 86.3& 96.1& 88.3& 86.5& 72.5& 70.6& 85.8& 87.2& 68.3& 76.4 & 82.6& 72.4& 89.4 \\
\vecoplugout & 88.3 & 67.4 & 87.4 & 88.5 & 86.7 & 95.9 & 89.0 & 87.8 & 75.1 & 70.9 & 86.2 & 88.9 & 67.5 & 76.2 & 82.9 & 72.9 & 89.9 \\
\midrule
\multicolumn{17}{l}{\textit{\textbf{Translate-Train-All}}} \\
\vecoplugin & 92.5 & 73.7 & 93.4 & 91.8 & 90.4 & 95.2 & 91.3 & 90.6 & 79.1 & 79.8 & 89.5 & 91.4 & 79.1 & 80.6 & 88.4 & 74.8 & 91.8 \\
\midrule
& ja & kk & ko & mr & nl & pt & ru & ta & te & th & tl & tr & ur & vi & yo & zh & \textbf{Avg.} \\
\midrule
\multicolumn{17}{l}{\textit{\textbf{Cross-lingual Transfer}}}\\
XLM-R & 15.9 & 78.1 & 53.9 & 80.8 & 89.5 & 87.6 & 89.5 & 65.2 & 86.6 & 47.2 & 92.2 & 76.3 & 70.3 & 56.8 & 24.6 & 25.7 & 73.8 \\
\vecoplugout & 31.4 & 79.3 & 53.1 & 84.3 & 89.8 & 88.3 & 90.2 & 64.3 & 85.8 & 48.0 & 93.7 & 77.2 & 69.2 & 58.1 & 26.2 & 39.4 & 75.1 \\
\midrule
\multicolumn{17}{l}{\textit{\textbf{Translate-Train-All}}} \\
\vecoplugin & 45.1 & 78.0 & 63.7 & 84.5 & 92.7 & 90.1 & 92.6 & 72.6 & 88.5 & 55.2 & 88.8 & 76.8 & 75.0 & 70.5 & 24.3 & 63.0 & 79.8 \\
\bottomrule
\end{tabular}
}
\caption{POS results (Accuracy) for each language.}
\label{tbl:pos}
\vspace{-3mm}
\end{table*}

\begin{table*}[!h]
\centering
\resizebox{\linewidth}{!}{
\begin{tabular}{lcccccccccccccccccccc}
\toprule
Model & en & af & ar & bg & bn & de & el & es & et & eu & fa & fi & fr & he & hi & hu & id & it & ja & jv \\
\midrule
\multicolumn{21}{l}{\textit{\textbf{Cross-lingual Transfer}}}\\
XLM-R & 84.7 & 78.9 & 53.0 & 81.4 & 78.8 & 78.8 & 79.5 & 79.6 & 79.1 & 60.9 & 61.9 & 79.2 & 80.5 & 56.8 & 73.0 & 79.8 & 53.0 & 81.3 & 23.2 & 62.5 \\
\vecoplugout & 83.8 & 77.5 & 48.2 & 83.9 & 77.2 & 79.4 & 79.3 & 75.4 & 80.4 & 68.3 & 68.2 & 80.6 & 80.1 & 55.0 & 71.0 & 80.9 & 52.9 & 81.7 & 19.4 & 63.2  \\
\midrule
\multicolumn{21}{l}{\textit{\textbf{Translate-Train-All}}} \\
\vecoplugin & 80.7 & 82.5 & 66.4 & 84.1 & 78.4 & 82.2 & 82.4 & 79.7 & 84.7 & 78.2 & 68.8 & 84.9 & 79.1 & 69.7 & 76.6 & 85.1 & 77.3 & 83.8 & 21.3 & 70.3 \\
\midrule
& ka & kk & ko & ml & mr & ms & my & nl & pt & ru & sw & ta & te & th & tl & tr & ur & vi & yo & zh \\
\midrule
\multicolumn{21}{l}{\textit{\textbf{Cross-lingual Transfer}}}\\
XLM-R & 71.6 & 56.2 & 60.0 & 67.8 & 68.1 & 57.1 & 54.3 & 84.0 & 81.9 & 69.1 & 70.5 & 59.5 & 55.8 & 1.3 & 73.2 & 76.1 & 56.4 & 79.4 & 33.6 & 33.1 \\
\vecoplugout & 67.1 & 51.2 & 59.9 & 63.4 & 65.0 & 70.0 & 56.1 & 83.4 & 83.1 & 71.3 & 70.5 & 60.5 & 56.2 & 1.4 & 71.3 & 80.4 & 69.3 & 76.0 & 37.4 & 29.1 \\
\midrule
\multicolumn{21}{l}{\textit{\textbf{Translate-Train-All}}} \\
\vecoplugin & 77.0 & 67.2 & 71.0 & 73.3 & 74.1 & 71.8 & 63.8 & 85.5 & 80.8 & 72.8 & 77.0 & 69.1 & 67.5 & 2.6 & 74.0 & 85.2 & 71.5 & 76.4 & 32.8 & 31.0 \\
\bottomrule
\end{tabular}
}
\caption{NER results (F1) for each language.}
\label{tbl:ner}
\vspace{-3mm}
\end{table*}

\begin{table*}[!h]
\centering
\resizebox{\linewidth}{!}{
\begin{tabular}{lcccccccccccc}
\toprule
Model & en & ar & de & el & es & hi & ru & th & tr & vi & zh & \textbf{Avg.} \\
\midrule
\multicolumn{13}{l}{\textit{\textbf{Cross-lingual Transfer}}}\\
XLM-R & 86.5 / 75.7 &  68.6 / 49.0 &  80.4 / 63.4 &  79.8 / 61.7 & 82.0 / 63.9 &  76.7 / 59.7 &  80.1 / 64.3 &  74.2 / 62.8 &  75.9 / 59.3 &  79.1 / 59.0 &  59.3 / 50.0 &  76.6 / 60.8 \\
\vecoplugout & 87.6 / 76.5 & 73.6 / 56.1 & 79.8 / 62.2 & 79.6 / 61.6 & 81.2 / 61.6 & 74.7 / 57.6 & 78.7 / 62.1 & 72.8 / 60.6 & 75.1 / 58.3 & 79.0 / 59.8 & 69.2 / 59.2 & 77.3 / 61.8 \\
\midrule
\multicolumn{13}{l}{\textit{\textbf{Translate-Train-All}}} \\
\vecoplugout & 88.3/77.9 & 76.9/61.1 & 80.5/64.6 & 81.5/64.1 & 84.2/66.8 & 78.8/62.5 & 80.2/66.1 & 77.0/70.4 & 77.8/62.2 & 82.5/63.7 & 71.6/69.4 & 79.9/66.3 \\
\vecoplugin & 90.2/79.5 & 81.8/66.4 & 85.4/69.8 & 85.3/69.0 & 87.2/70.8 & 83.7/67.9 & 85.6/71.6 & 80.0/74.7 & 82.4/68.6 & 85.8/68.3 & 74.9/73.1 & 83.9/70.9 \\
\bottomrule
\end{tabular}
}
\caption{XQuAD results (F1 / EM) for each language.}
\label{tbl:XQuAD}
\end{table*}

\begin{table*}[!h]
\centering
\resizebox{\linewidth}{!}{
\begin{tabular}{lcccccccc}
\toprule
Model & en & ar & de & es & hi & vi & zh & \textbf{Avg.} \\
\midrule
\multicolumn{9}{l}{\textit{\textbf{Cross-lingual Transfer}}}\\
XLM-R & 83.5 / 70.6 &  66.6 / 47.1 &  70.1 / 54.9 &  74.1 / 56.6 &  70.6 / 53.1 &  74.0 / 52.9 &  62.1 / 37.0 & 71.6 / 53.2\\
\vecoplugout & 83.6 / 70.5 & 65.0 / 44.6 & 69.8 / 54.6 & 73.8 / 55.6 & 69.1 / 51.4 & 73.1 / 51.8 & 67.3 / 43.6 & 71.7 / 53.2 \\
\midrule
\multicolumn{9}{l}{\textit{\textbf{Translate-Train-All}}} \\
\vecoplugout & 84.1/71.3 & 67.8/47.1 & 70.7/55.8 & 74.6/56.6 & 71.1/53.4 & 74.8/54.4 & 68.8/45.8 & 73.1/54.9  \\
\vecoplugin & 87.5/75.5 & 72.3/52.1 & 75.7/61.1 & 78.8/61.6 & 76.6/58.6 & 79.3/59.1 & 72.1/46.8 & 77.5/59.3 \\
\bottomrule
\end{tabular}
}
\caption{MLQA results (F1 / EM) for each language.}
\label{tbl:MLQA}
\end{table*}

\begin{table*}[!h]
\centering
\resizebox{\linewidth}{!}{
\begin{tabular}{lcccccccccc}
\toprule
Model &  en & ar & bn & fi & id & ko & ru & sw & te & \textbf{Avg.} \\
\midrule
\multicolumn{11}{l}{\textit{\textbf{Cross-lingual Transfer}}}\\
XLM-R & 71.5 / 56.8 &  67.6 / 40.4 &  64.0 / 47.8 &  70.5 / 53.2 &  77.4 / 61.9 &  31.9 / 10.9 &  67.0 / 42.1 &  66.1 / 48.1 &  70.1 / 43.6 &  65.1 / 45.0 \\
\vecoplugout & 71.3 / 58.2 & 73.1 / 52.8 & 58.9 / 42.5 & 70.9 / 55.1 & 77.2 / 60.0 & 54.2 / 39.9 & 66.1 / 37.6 & 65.8 / 45.7 & 70.6 / 50.7 & 67.6 / 49.1 \\
\midrule
\multicolumn{11}{l}{\textit{\textbf{Translate-Train-All}}} \\
\vecoplugout & 77.2/64.8 & 77.0/57.5 & 72.2/56.6 & 76.6/59.3 & 80.0/64.4 & 63.4/52.2 & 72.8/50.5 & 79.4/67.1 & 76.0/58.0 & 75.0/58.9 \\
\vecoplugin & 79.4/65.2 & 80.1/60.9 & 80.8/68.1 & 81.6/65.5 & 84.3/69.7 & 65.4/50.4 & 77.8/55.8 & 83.7/74.1 & 81.0/63.4 & 79.4/63.7 \\
\bottomrule
\end{tabular}
}
\caption{TyDiQA-GolP results (F1 / EM) for each language.}
\label{tbl:TyDiQA}
\vspace{-3mm}
\end{table*}

\begin{table*}[!h]
\centering
\label{tab:tatoeba_results}
\resizebox{\textwidth}{!}{
\begin{tabular}{lcccccccccccccccccc}
\toprule
Model & af  & ar  & bg  & bn  & de  & el  & es  & et  & eu  & fa  & fi  & fr  & he  & hi  & hu  & id  & it  & ja  \\
\midrule
\multicolumn{19}{l}{\textit{\textbf{Cross-lingual Transfer}}}\\
XLM-R  & 58.2  & 47.5  & 71.6  & 43  & 88.8 & 61.8 & 75.7 & 52.2 & 35.8 & 70.5 & 71.6 & 73.7 & 66.4 & 72.2 & 65.4 & 77  & 68.3 & 60.6 \\
\vecoplugout & 48.2 & 70.9 & 86.7 & 57.7 & 97.5 & 81.5 & 94.8 & 89.7 & 62.9 & 82.1 & 87.9 & 88.8 & 74.7 & 80.7 & 87.6 & 89.6 & 89.2 & 83.2 \\
\midrule
\multicolumn{19}{l}{\textit{\textbf{Translate-Train-All}}} \\
\vecoplugout & 80.9 & 85.1 & 91.3 & 78.1 & 98.5 & 89.5 & 97.4 & 94.8 & 79.8 & 93.1 & 95.4 & 93.7 & 85.8 & 94.2 & 93.8 & 93.0 & 92.2 & 92.8 \\
\vecoplugin & 88.5 & 88.7 & 91.5 & 84.2 & 98.9 & 91.5 & 97.9 & 96.4 & 85.8 & 95.3 & 95.9 & 95.6 & 89.6 & 97.0 & 95.1 & 94.2 & 94.1 & 94.0 \\
\midrule 
& jv  & ka  & kk  & ko  & ml  & mr  & nl  & pt  & ru  & sw  & ta  & te  & th  & tl  & tr  & ur  & vi  & zh  \\
\midrule
\multicolumn{19}{l}{\textit{\textbf{Cross-lingual Transfer}}}\\
XLM-R  & 14.1  & 52.1 & 48.5 & 61.4 & 65.4 & 56.8 & 80.8 & 82.2 & 74.1 & 20.3 & 26.4 & 35.9 & 29.4  & 36.7 & 65.7 & 24.3  & 74.7 & 68.3  \\
\vecoplugout & 17.6 & 58.5 & 53.9 & 75.3 & 80.1 & 64.2 & 94.4 & 92.8 & 88.6 & 37.4 & 61.9 & 65.8 & 84.5 & 52.5 & 89.3 & 64.3 & 85.8 & 82.7 \\
\midrule
\multicolumn{19}{l}{\textit{\textbf{Translate-Train-All}}} \\
\vecoplugout & 35.1 & 83.0 & 74.1 & 88.7 & 94.8 & 82.5 & 95.9 & 94.6 & 92.2 & 69.7 & 82.4 & 91.0 & 94.7 & 73.0 & 95.2 & 63.8 & 95.1 & 93.9 \\
\vecoplugin & 49.3 & 86.6 & 83.7 & 91.2 & 97.1 & 87.9 & 97.6 & 96.1 & 93.8 & 82.6 & 88.9 & 95.3 & 95.1 & 79.8 & 97.6 & 91.4 & 97.2 & 95.2 \\
\bottomrule
\end{tabular}}
\caption{Tatoeba results (Accuracy) for each language}
\end{table*}

\end{document}